%% file: main.tex
\PassOptionsToPackage{table,xcdraw,x11names}{xcolor}
\documentclass[10pt,twocolumn,letterpaper]{article}

\usepackage[pagenumbers]{cvpr} %

\input{preamble}

\usepackage[dvipsnames]{xcolor}
\definecolor{cvprblue}{rgb}{0.21,0.49,0.74}

\usepackage[pagebackref,breaklinks,colorlinks,allcolors=cvprblue]{hyperref}
\usepackage[normalem]{ulem}
\usepackage{tabulary}

\useunder{\uline}{\ul}{}

\def\paperID{*****} %
\def\confName{CVPR}
\def\confYear{2026}

\title{ImLoc: Revisiting Visual Localization with Image-based Representation}

\author{
    Xudong Jiang$^1$
    \quad
    Fangjinhua Wang$^1$\footnotemark[1]
        \quad
        Silvano Galliani$^2$
        \quad
        Christoph Vogel$^2$
        \quad
        Marc Pollefeys$^{1,2}$\\
        $^1$Department of Computer Science, ETH Zurich\\
        $^2$Microsoft Spatial AI Lab, Zurich}

\begin{document}
\maketitle
{
    \renewcommand{\thefootnote}
        {\fnsymbol{footnote}}
        \footnotetext[1]{Corresponding author.}
}
\input{sec/0_abstract}

\input{sec/1_intro}

\input{sec/2_relatedwork.tex}

\input{sec/3_method}

\input{sec/4_experiments}

\input{sec/5_conclusion}
\clearpage

{
    \small
    \bibliographystyle{ieeenat_fullname}
    \bibliography{main}
}

\input{sec/X_suppl}

\end{document}

%% file: preamble.tex
\newcommand\customparagraph[1]{\vspace{0.4em}\noindent\textbf{#1.}}
\newcommand\customquestion[1]{\vspace{0.0em}\noindent\textbf{#1?}}
\newcommand{\mycomment}[1]{}
\newcommand{\Tab}[1]{Tab.~#1}
\newcommand{\Fig}[1]{Fig.~#1}

\usepackage{multirow}

%% file: sec/0_abstract.tex
\begin{abstract}
Existing visual localization methods are typically either 2D image-based, which are easy to build and maintain but limited in effective geometric reasoning, or 3D structure-based, which achieve high accuracy but require a centralized reconstruction and are difficult to update. 
In this work, we revisit visual localization with a 2D image-based representation and propose to augment each image with estimated depth maps to capture the geometric structure. 
Supported by the effective use of dense matchers, this representation is not only easy to build and maintain, 
but achieves highest accuracy in challenging conditions. 
With compact compression and a GPU-accelerated LO-RANSAC implementation, the whole pipeline is efficient in both storage and computation and allows for a flexible trade-off between accuracy and highest memory efficiency. 
Our method achieves a new state-of-the-art accuracy on various standard benchmarks and outperforms existing memory-efficient methods at comparable map sizes. 
Code will be available at \url{https://github.com/cvg/Hierarchical-Localization}
\end{abstract}

%% file: sec/1_intro.tex
\section{Introduction}
\label{sec:intro}

Visual localization describes the task of estimating the camera position and orientation for a query image in a scene defined by a set of database images with known poses. 
It is a key challenge in applications like robotics, autonomous driving, and Augmented / Virtual Reality. %
Currently, the approaches for visual localization can be divided into two main categories: \emph{2D image-based} and \emph{3D structure-based} localization. 

\begin{figure}[t]
    \centering

\begin{minipage}[t!]{0.33\linewidth}
        \centering
        \includegraphics[width=\linewidth]{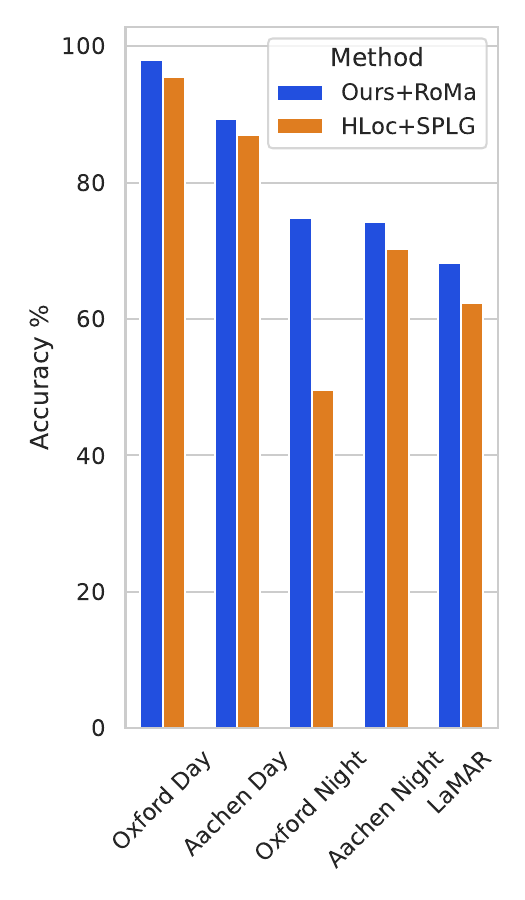}
    \end{minipage}%
    \begin{minipage}[t!]{0.66\linewidth}
        \centering
        \includegraphics[width=\linewidth]{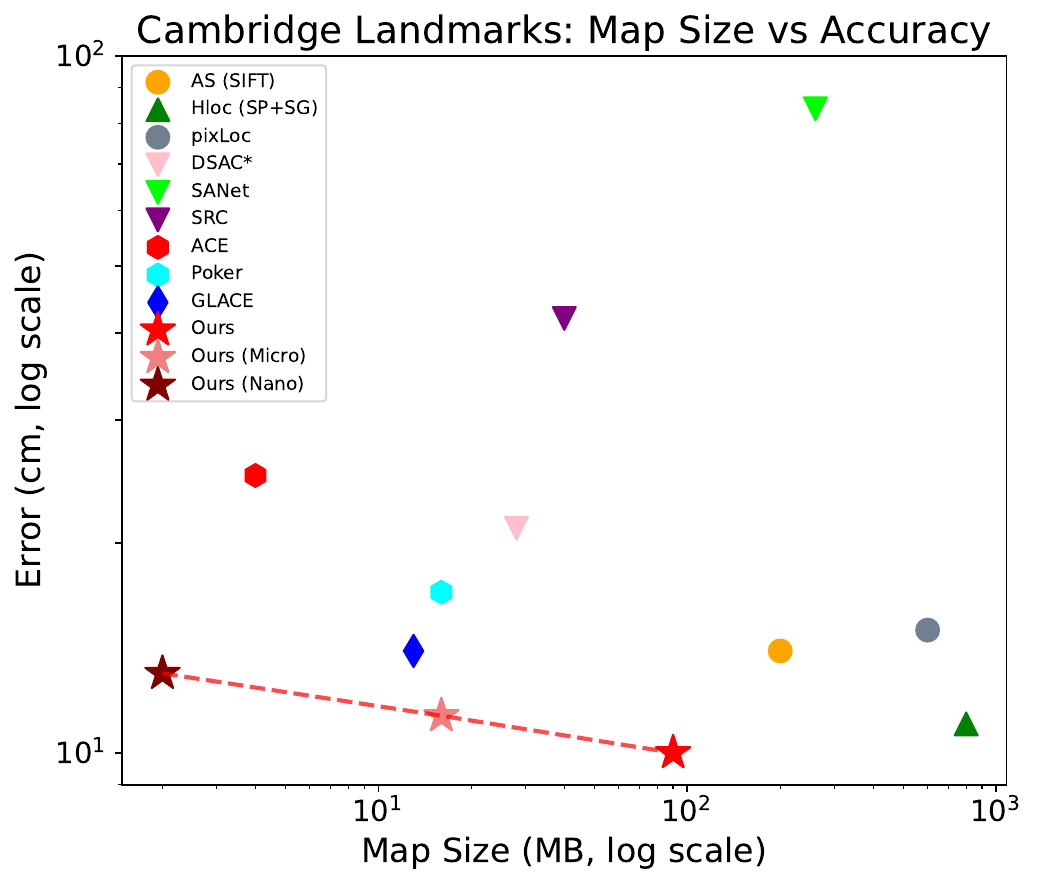}
        
    \end{minipage}%
\vspace{-0.45cm}
    \caption{\emph{Left:} \emph{ImLoc} achieves state-of-the-art results on a multitude of datasets: LaMAR~\cite{sarlin2022lamar}, Aachen Day and Night 1.1~\cite{sattler2012bmvc, sattler2018cvpr}, Oxford Day and Night~\cite{wang2025seeing}, Cambridge Landmarks~\cite{kendall2015posenet}, surpassing the previous gold standard HLoc~\cite{sarlin2019coarse}. \emph{Right:} \emph{ImLoc} (\({\color{Maroon} \star}, {\color{Salmon} \star}, {\color{Red} \star}\)) allows a trade-off between accuracy and memory efficiency and maintains state-of-the-art accuracy at various compression levels. }
    \label{fig:teaser}
    \vspace{-0.2cm}
\end{figure}

\begin{figure*}[t]
    \centering
    \includegraphics[trim={0 3 0 2}, clip, width=1.0\linewidth]{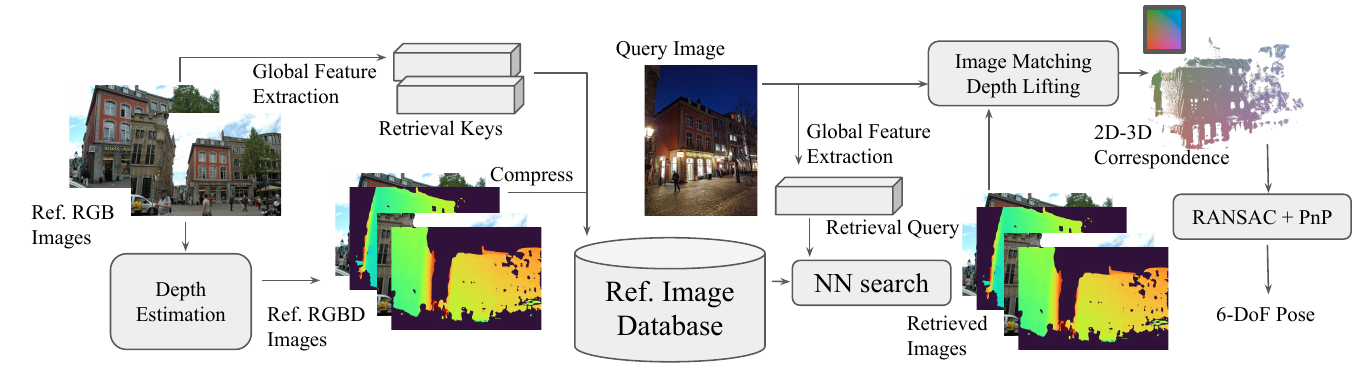}
    \caption{\textbf{Illustration of our localization pipeline.} During mapping, we store RGB and depth images along with camera poses, intrinsics, and retrieval features. For localization, we run dense image matching between the query and the top-K retrieved database images and establish 2D-3D correspondences using the precomputed depth maps. The camera pose is estimated with PnP+RANSAC. Please refer to Section~\ref{sec:pipeline} for details.}
    \vspace{-0.4cm}
    \label{fig:pipeline}
\end{figure*}

2D image-based localization methods represent a scene as a database of calibrated images. 
Given a query image, after a set of relevant images is selected from the database by image retrieval~\cite{arandjelovic16netvlad}, the relative pose between query and retrieved images~\cite{sattler2017large,panek2025guide} is computed and returned. 
In contrast, 3D structure-based localization methods store the 3D geometry of the scene, either explicitly as point cloud, mesh, 3D Gaussian Splatting (3DGS)~\cite{kerbl20233d}, NeRF~\cite{mildenhall2020nerf}, or implicitly in the weights of a neural network~\cite{brachmann2023ace,glace2024cvpr}. They establish matches between 2D pixels in the query image and 3D points in the scene, which are used to predict camera pose with a Perspective-n-Point (PnP) solver~\cite{haralick1994review,bujnak2008general}. 

The common view of the research community is that there exists a trade-off between both representations.
2D image-based methods appear more scalable and flexible because they do not need to compute and store globally consistent 3D geometry.
In contrast, 3D structure-based methods~\cite{sarlin2019coarse} commonly employ triangulation and store a centralized point cloud to represent the scene, which makes the representation less flexible. %
Consequently, 3D representations are often limited to static reconstruction, and cannot handle dynamics and scene changes. 
Another downside of advanced forms of 3D representations like mesh, 3DGS and NeRF is that they cannot accurately represent the scene with limited model capacity. 
To avoid this drawback and also to speed up query localization, sparse methods~\cite{sarlin2019coarse} reduce map size through keypoint selection or map compression \cite{brachmann2023ace,compression2019cvpr}
One major disadvantage of 2D image-based methods, however, is that their accuracy is generally worse than that of 3D structure-based methods, as extensive geometric reasoning typically leads to better performance~\cite{panek2025guide}. 

In this paper, we argue in favor of a lightweight, simple and flexible 2D image-based representation. 
We introduce \emph{ImLoc} to combine the advantages of image-based and structure-based representations.
Specifically, we avoid committing to a globally consistent 3D structure and instead store 3D structure as 2D image-based representation,~\ie, we predict and store 2D depth maps along with RGB images, intrinsics, and extrinsics.
Retaining the original geometric source without premature abstraction, allows us to leverage the latest advances in depth estimation and dense matching to achieve unprecedented accuracy and robustness. 
This representation serves as a unified interface for sparse~\cite{sarlin2020superglue,lindenberger2023lightglue} and dense matching~\cite{edstedt2023roma}, feed-forward, and refinement models. %
It allows us to easily trade-off compression for accuracy when storing the `map' and to switch between models without re-constructing the 3D structure, while also enabling trade-offs between accuracy and efficiency at test time. 
In contrast to sparse 3D structure-based methods, we postpone the decision on which points can be selected as correspondences until after the matching stage, which leads to improved accuracy. 
An efficient GPU accelerated LO-RANSAC allows to effectively utilize the dense correspondences for estimating the pose. 
As shown in Fig.~\ref{fig:teaser} (\emph{right}), \emph{ImLoc} achieves state-of-the-art results on several large-scale benchmarks~\cite{sattler2012bmvc,kendall2015posenet,sarlin2022lamar,wang2025seeing} with a reasonable memory footprint. 
Furthermore (Fig.~\ref{fig:teaser}, \emph{left}), \emph{ImLoc} attains state-of-the-art accuracy at various desired compression levels (measured on Cambridge Landmarks~\cite{kendall2015posenet}).

Our contributions are summarized as follows: 
\begin{itemize}
\item Following Occam’s Razor, we propose \emph{ImLoc}, a simple and scalable localization pipeline that provably generalizes well to various datasets.
\item By not committing to an explicit and consistent global 3D structure, \emph{ImLoc} provides an attractive level of simplicity and flexibility. %
\item While \emph{ImLoc} is already highly efficient in storage and competitive in run-time, it offers additional options to further trade-off memory efficiency and runtime for accuracy, in both the mapping (\eg image compression) and the localization stage (\eg density of correspondences).
\end{itemize}

%% file: sec/2_relatedwork.tex
\section{Related Work}
\label{sec:relatedwork}

\subsection{Structure-based Geometric Modeling}
In structure based geometric modeling, the visual localization problem is formalized as a probabilistic model, and %
factored into independent observations that pose geometric constraints -- those constraints often are given as 2D-3D point correspondences. 
Here, each term contributes via a robustified distribution of the reprojection error. %
The dominant representation are sparse (SfM) 3D point clouds, descriptor matching to establish correspondences and RANSAC to facilitate robust pose estimation.

\subsection{Coarse-to-fine Localization}
To efficiently localize across different scales,~\eg, city-scale scenes, a common strategy is to use a coarse-to-fine pipeline~\cite{sarlin2019coarse}. %
Specifically, the coarse part is achieved by image retrieval~\cite{arandjelovic16netvlad,radenovic2018fine,zhu2023r2former} and the fine by feature matching between the retrieved database images and the query image~\cite{sarlin2019coarse,sarlin2020superglue,lindenberger2023lightglue}. 
In this work, we stick to the coarse stage of common pipelines (~\ie, image retrieval), but revisit the fine localization part. 

\subsection{Scene Representations for Visual Localization}

\customparagraph{SfM Point Cloud}
Most state-of-the-art visual localization methods~\cite{sattler2016efficient,sarlin2019coarse,sarlin2020superglue,detone2018superpoint} follow structured geometric modeling and represent the scene as a point cloud obtained via Structure-from-Motion (SfM) to achieve high accuracy and robustness. 
Following the classical SfM pipeline~\cite{schoenberger2016sfm}, they detect keypoints in the reference images, extract local descriptors, match them across images to establish 2D-2D correspondences, and triangulate them to obtain the corresponding 3D points. 
Usually, the sparse matcher for SfM is also used during localization, and the local descriptors are stored together with the 3D points to facilitate 2D-3D matching. 
To reduce storage in large scenes, various compression techniques have been proposed, including point cloud sparsification~\cite{li2010location,camposeco2019hybrid,yang2022scenesqueezer} and descriptor compression \cite{dong2023learning,ke2004pca,yang2022scenesqueezer,lynen2015get,Laskar2024dpqed}. 
In addition, several studies~\cite{zhou2022gomatch,wang2024dgc,panek2022meshloc} avoid storing descriptors and directly match against geometric representations. However, these methods must trade off accuracy for memory efficiency. 

\customparagraph{Image-based Representation}
Image-based representation~\cite{sattler2017large,panek2025guide} propose to store only images and no explicit 3D geometry in their map. This simple representation can naturally handle dynamic scene changes by adding or removing images. 
The pose of a query can be approximated by retrieving the most similar reference image and using its pose~\cite{sattler2017large}, or by interpolating the poses of the top-N retrieved images~\cite{sattler2019understanding,panek2025guide}. 
A recent survey~\cite{panek2025guide} shows that localization accuracy can be improved by utilizing the local geometric structure determined from the constraints of the retrieved images. 
For example, ~\cite{sattler2017large} establish a locally consistent SfM model on the fly at query time and show that pose estimation with a local, structure-based model performs better than simple pose approximation. %
However, we note that structural information does not need to be recomputed repeatedly at test time and the representation does not need to be even locally consistent. %
Instead, it can be precomputed and stored in a 2D depth map for each view. 
Given the cost of storing the images, we show that storing additional geometric information like depth maps does not introduce a significant overhead, but explicitly facilitates geometric reasoning at query time. 
Similarly, InLoc~\cite{taira2018inloc} explored RGB-D panoramas for indoor localization. 
However, it requires specialized sensors during mapping and are limited to indoor environments. 
In this work, we propose to augment each image with \emph{pre-computed} depth maps to improve geometric reasoning, and achieve the high accuracy of structure-based modeling, while maintaining the flexibility and easy maintenance of image-based representations.

\customparagraph{Scene Coordinate Regression and Pose Regression} 
These methods do not store an explicit map of the scene, but directly model the discriminative relationship, usually in the form of a neural network that implicitly encodes the geometric structure of the scene. 
\newline
Trained end-to-end, Pose Regression (PR) methods~\cite{kendall2015posenet,Kendall2017GeometricLF,Brahmbhatt2018mapnet,Shavit2021MStransformer,turkoglu2021visual,WinkelbauerICRA21,zhou2020essnet,naseer2017deep} 
directly regress an absolute or relative pose for a query from the input image in a feed-forward manner. %
Absolute-PR~\cite{kendall2015posenet,Kendall2017GeometricLF,naseer2017deep,Walch2017lstm} typically struggles with generalization and does not scale well with limited network capacity~\cite{taira2018inloc}. 
Relative-PR~\cite{balntas2018relocnet,ding2019camnet,zhou2020essnet} is scene-agnostic and regresses a camera pose relative to database images, but is often limited in accuracy. 
\newline
Scene Coordinate Regression (SCR) methods~\cite{brachmann2016,Cavallari2019cascade,brachmann2017dsac,brachmann2023ace,wang2024hscnet++,glace2024cvpr,jiang2025r} 
instead first establish 2D-3D correspondences between the query image and the scene, by regressing the corresponding 3D coordinate for each 2D pixel. 
SCR~\cite{brachmann2023ace} is highly efficient in storing the map. 
Recently, several approaches improve the scalability and performance of SCR in large scenes~\cite{brachmann2019esac,tang2023neumap,li2020hscnet,wang2024hscnet++,glace2024cvpr,jiang2025r}.
However, these methods still encounter limitations under challenging conditions %
and have an accuracy gap compared to top feature matching methods~\cite{wang2025seeing}. 
Precomputing an implicit representation for a scene, it is difficult for both PR and SCR to handle scene changes and dynamics.

\customparagraph{Novel View Synthesis}
Approaches to utilize novel view synthesis (NVS)~\cite{mildenhall2020nerf,kerbl20233d} for visual localization include NeRF-~\cite{liu2023nerf}, 
Gaussian splatting-~\cite{botashev2024gsloc}, and mesh-based~\cite{panek2022meshloc,trivigno2024unreasonable} methods. 
Localization can be performed by a render-and-compare framework~\cite{labbe2022megapose,li2018deepim}, where the pose is found by aligning the rendered image with the query image~\cite{chen2024neural, yen2021inerf,lin2023parallel,trivigno2024unreasonable}, 
or by combining NVS with structure-based modeling~\cite{zhou2024nerfmatch,germain2022feature,moreau2023crossfire,liu2023nerf,panek2022meshloc,liu2025gscpr} to establish better 2D-3D correspondences. 
NVS-based methods 
are challenged by illumination changes during mapping and testing and cannot handle dynamics and scene changes effectively due to the need to build a globally consistent model.

%% file: sec/3_method.tex
\section{Method}
\label{sec:method}

\input{table/oxford_day_night.tex}

We consider \emph{coarse-to-fine localization} and \emph{structure-based geometric modeling} as key ingredients for scalable, robust visual localization. 
Based on these core principles we try to follow Occam's Razor to build a minimalistic image-based visual localization pipeline by avoiding too many additional assumptions to maximize its generalization capabilities. 

\subsection{Pipeline}\label{sec:pipeline}
Our image-based localization pipeline is shown in Fig.~\ref{fig:pipeline}.

\customparagraph{Scene Representation}
We store precomputed image-retrieval features~\cite{berton2025megaloc,berton2023eigenplaces,arandjelovic16netvlad} for coarse localization. 
For fine localization, we store the original RGB images, which enable 2D-2D matching. 
In addition, we predict and store (dense) depth maps together with poses and camera intrinsics. 
This provides minimal but sufficient information that enables us to flexibly lift 2D-2D matches to 2D-3D correspondences for structure-based pose estimation, 
without prematurely committing to sparse (key-point) locations. 

\customparagraph{Mapping Pipeline for Building the Database}
Our simplistic representation allows for a scalable and flexible mapping pipeline. RGB images with known poses and intrinsics can be processed independently to estimate depth maps and extract global retrieval features. We can store them together with posed RGB images in the database with optional compression to reduce storage. 
For depth estimation, we can flexibly use various depth models. 
When adding new images to the database, we can perform depth estimation using only the new images and/or retrieve existing database images that are potentially covisible. 

\customparagraph{Hierarchical Structured-based Localization}
Given a query image, we first extract its global feature~\cite{berton2025megaloc,berton2023eigenplaces,arandjelovic16netvlad} and retrieve the top-$K$ database images. 
Then we perform image matching between the query image and retrieved database images to establish 2D-2D correspondences, which are lifted to 2D-3D correspondences with the precomputed depth maps of the database images. 
Finally, we robustly estimate the camera pose by running PnP+RANSAC on the 2D-3D correspondences. 
By storing the depth densely, we retain the freedom to choose which and how many correspondences we want to use here. 

\subsection{Motivation}

In this section, we reflect on the design choices of our pipeline. %
Our aim is to maximize flexibility and accuracy while maintaining simplicity and scalability. 
\newline
\customquestion{Why retrieval}
Image retrieval enables scalable, coarse-to-fine localization by efficiently narrowing down candidate database images for further geometric reasoning. 
This hierarchical approach balances computational efficiency with localization accuracy.
\newline
\customquestion{Why posed RGB images}
Storing posed RGB images preserves the full visual information of the scene, ensuring compatibility with advances in image-based models, compression, and matching. 
Unlike premature abstraction to keypoints or descriptors, RGB images retain all information, providing an upper bound for localization accuracy and benefiting from modern compression methods, 
which can be more efficient than storing descriptors alone~\cite{panek2022meshloc}.
\newline
\customquestion{Why geometry in image-based representation (depth)}
Previous works~\cite{panek2025guide} show that geometric reasoning is important for accurate localization. 
Augmenting each image with a precomputed depth map enables effective geometric reasoning without requiring a globally consistent 3D model. 
Depth maps compactly encode 2D–3D correspondences per pixel, facilitating the use of dense matching methods during localization, while demanding only local geometric reasoning during mapping. 
We show that even dense depth can be efficiently estimated (or acquired by sensors) and stored in compressed form with little overhead. 
\newline
\customquestion{Why not NVS or globally consistent geometry}
Novel view synthesis (NVS) models like NeRF, Gaussian Splatting, or mesh-based approaches can render new views, but require building and maintaining a globally consistent 3D model, which is challenging in dynamic or sparsely observed scenes. 
These models are often less compact and more difficult to update than our image-based representation with depth. 
By avoiding a globally consistent map, our approach remains flexible, easy to maintain, and robust to scene changes.

\subsection{Implementation}
For the implementation of our image-based visual localization pipeline we utilize a robust dense image matching model, RoMa~\cite{edstedt2023roma}, for both mapping time depth estimation and query time 2D-2D matching. Our dense image-based representation allows to fully leverage the power of modern dense image matching models, and shows strong potential in robust and accurate visual localization. 

\customparagraph{Dense Image Matching}
Instead of detecting sparse keypoints and matching between them, dense image matching methods aim to find the matched position in another image for each pixel in the reference image. %
This provides a general formulation of matching without premature abstraction and quantization to keypoints. 
It avoids the challenges imposed by keypoint detection repeatability, and allows to leverage all the image information for matching. 
Recent advances in deep learning based dense matching models~\cite{edstedt2023roma} have shown strong robustness in challenging conditions. 
Naively matching all pixels can harm overall efficiency, especially for high-resolution images. 
We empirically find that setting the resolution as $560\!\times\!560$ for both depth map and RGB image achieves strong localization performance at affordable computational cost. 
Accordingly, we only need to store relatively low-resolution images in the database, which reduces storage requirements. 
We conjecture that low-resolution images retain most of the important information for localization or potentially other perception tasks, while high frequency details may be less important. 
In contrast, keypoint based methods often need high resolution images to ensure accurate keypoint localization. 

\customparagraph{Depth Estimation by Dense Matching}
Although there exist many multi-view depth estimation methods~\cite{yao2018mvsnet,wang2021patchmatchnet,wang2022itermvs,izquierdo2025mvsanywhere,wang2025lightweight}, we find their robustness limited in scenes with strong illumination changes. 
In contrast, we observe that dense matching models~\cite{edstedt2023roma} are usually trained on more diverse datasets~\cite{li2018megadepth} and generalize better. 
Therefore, we perform dense matching with RoMa~\cite{edstedt2023roma} to estimate depth maps, similar to triangulation. 
At first, we select covisible images from the database for each mapping image. 
The covisibility is estimated with the reference SfM reconstruction~\cite{schoenberger2016sfm}, or by retrieving images with similar global retrieval features~\cite{berton2025megaloc}. %
Then we perform dense RoMa matching to find for each pixel in the mapping image the corresponding pixels in the retrieved images and a confidence score. 
We filter the matches using a confidence threshold ($\geq0.05$) and triangulate the depth for each pixel with valid matches. 
Following best practice~\cite{schoenberger2016sfm}, we use robust estimation to handle outliers. 
As poses are known, there is only one degree of freedom, and a single match is enough to provide a depth hypothesis. 
We exhaustively try all matches and select the one with the most inliers for an angular error threshold of 2 degrees. 
Then we refine the depth by minimizing the sum of squared angular errors of inliers, weighted by RoMa matching confidence. 
Finally, we keep the depth estimates with more than 3 inliers. 
To maximize efficiency, we implement this dense triangulation on the GPU. 
Triangulation usually takes about 30ms per image on RTX 4090 GPU and the main bottleneck is the matching time of RoMa.

\customparagraph{Data Compression}
To further improve the compactness of our representation, we compress the RGB images and depth maps. For RGB images, we downsample them to $560\!\times\!560$ resolution with the LANCZOS filter~\cite{duchon1979lanczos}, and compress them with JPEG XL~\cite{alakuijala2019jpeg} at quality 90. 
For depth maps, we clip the depth in the range from $0.25m$ to $128m$, quantize the depth to log space with 256 levels, 
and finally compress the quantized depth map with JPEG XL~\cite{alakuijala2019jpeg} lossless compression. 

\customparagraph{Image Retrieval}
To allow for a fair comparison with other baselines, we use their image retrieval settings in the experiments. 
If we can easily run all the baselines, we use Megaloc~\cite{berton2025megaloc} for retrieval. 
Specifically, we use Megaloc~\cite{berton2025megaloc} for Oxford Day \& Night~\cite{wang2025seeing} and LaMAR~\cite{sarlin2022lamar}, NetVLAD~\cite{arandjelovic16netvlad} for Cambridge Landmarks~\cite{kendall2015posenet}, and EigenPlaces~\cite{berton2023eigenplaces} for Aachen Day-Night~\cite{sattler2018cvpr, sattler2012bmvc}. 

\customparagraph{Query Time Matching and 2D-3D Lifting}
Given a query image and the retrieved database images, we perform bidirectional dense matching with RoMa between the query image and each retrieved database image to establish 2D-2D correspondences. 
Matching from query to database, we bilinearly interpolate the depth at subpixel coordinates; otherwise we can just lookup the values. 
Matches with a valid depth and a confidence greater than $0.05$ form the final 2D-3D correspondences for pose estimation. 

\customparagraph{Pose Estimation}
To flexibly handle a large number of 2D-3D correspondences from dense matching, we implement a GPU accelerated LO-RANSAC~\cite{loransac} following poselib~\cite{poselib} for robust pose estimation. 
Since densely matched points are highly correlated spatially, we use uniform subsampling to limit the correspondences to a maximum of 10K for scoring pose hypotheses within RANSAC. 
We use all the inliers for the final refinement, using a robust Cauchy loss, weighted by the RoMa~\cite{edstedt2023roma} confidences. 
In detail: the CPU samples a batch of 1K minimal sets and generates hypotheses using the poselib P3P solver~\cite{poselib}. 
Then the GPU scores all hypotheses in parallel on the reduced correspondence set, using a truncated squared reprojection error weighted by RoMa confidence. 
When a new best hypothesis is found, we perform non-linear refinement of the truncated squared reprojection error on the downsampled correspondences. 
The RANSAC stops after 100K iterations or if the probability of missing the best model is below $10^{-4}$. 
For a final refinement with our GPU implementation, we employ all inliers in the full set of correspondences, using the robust Cauchy loss weighted by the RoMa confidence.

%% file: table/oxford_day_night.tex
{
\setlength{\tabcolsep}{3pt} %
\begin{table*}[th!]
  \resizebox{1.0\textwidth}{!}{%

    \begin{tabular}{l|lccccc|c}
      \toprule
      \multicolumn{7}{c}{\textbf{Visual Relocalization Results on \textit{Day} Queries}}                                                                                                                                                                                                                  \\ \midrule
                            &                                                             & \textbf{Bodleian Library}       & \textbf{H.B. Allen Centre}         & \textbf{Keble College}           & \textbf{Observatory Quarter}    & \textbf{Robotics Institute}     & \textbf{Average}                \\ \midrule
      \multirow{3}{*}{Hloc} & SP+LG~\cite{detone2018superpoint,lindenberger2023lightglue} & 96.64 / 98.78 / 99.16           & 98.73 / 99.37 / 99.37              & 95.06 / 97.91 / 98.48            & 95.75 / 96.70 / 96.70           & 91.02 / 92.52 / 93.02           & 95.44 / 97.06 / 97.35           \\
                            & Aliked+LG~\cite{zhao2023aliked, lindenberger2023lightglue}  & 97.56 / 99.16 / 99.54           & 99.37 / 99.37 / 99.37              & 96.39 / 98.86 / 99.62            & 95.52 / 96.46 / 96.46           & 88.28 / 90.27 / 91.02           & 95.42 / 96.82 / 97.00           \\
                            & RoMa~\cite{edstedt2023roma}(our)                            & 93.89 / 96.56 / 97.02           & 96.20 / 97.47 / 97.47              & 93.92 / 97.53 / 97.91            & 93.87 / 95.28 / 95.52           & 91.27 / 92.27 / 92.77           & 93.83 / 95.62 / 96.14           \\
      \midrule
      \multirow{3}{*}{SCR}  & ACE~\cite{brachmann2023ace}                                 & 0.00 / 0.00 /  0.99             & 0.63 / 8.86 / 31.65                & 0.57 / 3.80 / 22.24              & 0.24 / 8.02 / 25.24             & 0.00 / 2.24 / 11.72             & 0.29 / 4.58 / 18.37             \\
                            & GLACE~\cite{glace2024cvpr}                                  & 0.00 / 0.61 / 10.38             & 0.63 / 4.43 / 34.81                & 0.19 / 4.18 / 35.93              & 0.24 / 6.13 / 33.02             & 0.00 / 0.75 / 29.43             & 0.21 / 3.22 / 28.71             \\
                            & R-SCoRe~\cite{jiang2025r}                                   & 47.71 / 68.32 / 79.62           & 50.00 / 64.56 / 73.42              & 60.46 / 75.10 / 85.74            & 45.52 / 58.02 / 71.23           & 5.99 / 12.47 / 18.20            & 41.94 / 55.69 / 65.64           \\
      \midrule
      Ours                  & RoMa                                                        & \textbf{98.40 / 99.69 / 99.77 } & \textbf{100.00 / 100.00 / 100.00 } & \textbf{98.10 / 99.62 / 100.00 } & \textbf{98.11 / 98.58 / 98.58 } & \textbf{95.01 / 96.26 / 96.51 } & \textbf{97.92 / 98.83 / 98.97 } \\
      \midrule
      \multicolumn{7}{c}{\textbf{Visual Relocalization Results on \textit{Night} Queries}}                                                                                                                                                                                                                \\ \midrule
                            &                                                             & \textbf{Bodleian Library}       & \textbf{H.B. Allen Centre}         & \textbf{Keble College}           & \textbf{Observatory Quarter}    & \textbf{Robotics Institute}                                       \\ \midrule
      \multirow{4}{*}{Hloc} & SP+LG~\cite{detone2018superpoint,lindenberger2023lightglue} & 46.08 / 57.15 / 63.65           & 53.67 / 63.92 / 70.16              & 17.21 / 23.23 / 28.06            & 55.89 / 62.59 / 67.36           & 76.19 / 78.42 / 79.13           & 49.61 / 57.06 / 61.67           \\
                            & Aliked+LG~\cite{zhao2023aliked, lindenberger2023lightglue}  & 59.53 / 69.33 / 73.76           & 65.03 / 73.72 / 77.95              & 24.62 / 32.23 / 38.91            & 62.07 / 72.50 / 78.24           & 76.49 / 79.84 / 81.16           & 57.55 / 65.52 / 69.80           \\
                            & RoMa~\cite{edstedt2023roma}(paper~\cite{wang2025seeing})    & 70.25 / 79.09 / 82.22           & 66.37 / 81.51 / 87.31              & 31.50 / 42.22 / 51.82            & 72.06 / 80.92 / 84.50           & 78.22 / 82.27 / 83.69           & 63.68 / 73.20 / 77.91           \\
                            & RoMa~\cite{edstedt2023roma}(our)                            & 74.93 / 83.32 / 85.73           & 81.74 / 89.09 / 93.54              & 35.47 / 45.20 / 53.28            & 77.20 / 84.50 / 86.66           & 83.49 / 85.82 / 86.73           & 70.57 / 77.59 / 81.19           \\

      \midrule
      \multirow{3}{*}{SCR}  & ACE~\cite{brachmann2023ace}                                 & 0.00 / 0.00 / 0.00              & 0.00 / 0.00 / 0.00                 & 0.00 / 0.00 / 0.00               & 0.00 / 0.00 / 0.00              & 0.10 / 0.10 / 0.91              & 0.02 / 0.02 / 0.18              \\
                            & GLACE~\cite{glace2024cvpr}                                  & 0.00 / 0.00 / 0.03              & 0.00 / 0.00 / 0.00                 & 0.00 / 0.00 / 0.99               & 0.00 / 0.00 / 0.00              & 0.00 / 0.00 / 8.21              & 0.00 / 0.00 / 1.85              \\
                            & R-SCoRe~\cite{jiang2025r}                                   & 2.72 / 7.57 / 13.10             & 5.57 / 11.58 / 23.61               & 0.20 / 0.99 / 1.92               & 3.06 / 7.75 / 13.34             & 2.13 / 6.08 / 9.52              & 2.74 / 6.79 / 12.30             \\
      \midrule
      Ours                  & RoMa                                                        & \textbf{79.85 / 85.49 / 87.55 } & \textbf{85.97 / 93.76 / 95.77 }    & \textbf{37.39 / 46.33 / 53.54 }  & \textbf{83.31 / 87.56 / 88.52 } & \textbf{87.23 / 89.46 / 90.07 } & \textbf{74.75 / 80.52 / 83.09 } \\
      \bottomrule
    \end{tabular}
  }
  \caption{\textbf{Visual Relocalization Results on Oxford \textit{Day} and \textit{Night}~\cite{wang2025seeing}.} We report the percentage of query images correctly localized within three thresholds: (0.25m, 2\textdegree), (0.5m, 5\textdegree) and (1m, 10\textdegree). Results are shown for both Hloc with feature-matching and scene coordinate regression (SCR) approaches.
    For all feature-matching approaches, we use the top 50 images retrieved by Megaloc for matching.
  }\label{table:reloc_combined}
  \vspace{-0.2cm}
\end{table*}
}

%% file: sec/4_experiments.tex
\section{Experiments}
\label{sec:experiments}

\input{table/lamar}

\input{table/cambridge}

\input{table/aachenv11}
\subsection{Datasets}

To demonstrate that our pipeline generalizes well to different conditions, we evaluate it on many well known datasets that are popular for benchmarking localization pipelines. 

\noindent\textbf{Oxford Day \& Night}~\cite{wang2025seeing} is a recent large-scale egocentric dataset with challenging lighting conditions, including two sets of images to benchmark both day and night localization. It spans over 30 $km$ of recorded trajectories and covers an area of $40,000 m^2$. In total, the dataset comprises 5466 database images, 2819 daytime query images, and 7179 nighttime query images.

\noindent\textbf{LaMAR}~\cite{sarlin2022lamar} is a benchmark of large-scale scenes recorded with head-mounted and hand-held AR devices. It covers an area of $45,000 m^2$ and was acquired over one year. It is unique for its scale but also because it contains short-term appearance and structural changes due to moving people, weather, or day-night cycles, and long-term changes due to displaced furniture or construction work. The mapping set contains 97148 images and the phone query set contains 4477 images. %

\noindent\textbf{Cambridge Landmarks}~\cite{kendall2015posenet} is a large-scale outdoor dataset with RGB sequences of landmarks in Cambridge. It includes ground truth poses and a sparse 3D reconstruction generated via SfM. The data is split into 5365 mapping and 1918 query images.

\noindent\textbf{Aachen Day-Night}~\cite{sattler2012bmvc, sattler2018cvpr} is a city-scale dataset for outdoor visual localization, covering an area of approximately 6 $km^2$. It presents significant challenges due to varying illumination conditions, especially between day and night. The dataset contains 6697 reference and 1015 query images.

\subsection{Benchmark Performance}

\noindent\textbf{Oxford Day \& Night}. 
In \Tab\ref{table:reloc_combined} we compare \emph{ImLoc} to state-of-the-art (\cf\cite{wang2025seeing}) feature matching based~\cite{sarlin2019coarse} and to SCR~\cite{brachmann2023ace,glace2024cvpr,jiang2025r} methods. 
\emph{ImLoc} delivers the best accuracy under all conditions (day/night), in any scene and at any error threshold. 
The performance of our proposed pipeline is not only driven by the robustness and accuracy of the dense matcher, but it can only fully utilize the power of dense matching due to our dense geometric representation. 
Note that HLoc~\cite{sarlin2019coarse} equipped with dense RoMa~\cite{edstedt2023roma} features performs worse than with other sparse features~\cite{detone2018superpoint,lindenberger2023lightglue,sarlin2020superglue}, indicating that HLoc cannot exploit dense (RoMa) matching as well as \emph{ImLoc}. 
HLoc still needs to downsample and quantize the matches to make it work within reasonable resources. 
Instead, \emph{ImLoc} benefits from a denser set of correspondences and postpones correspondence selection to the post matching stage, while any sparse method commits to the point selection already at mapping time. 

\noindent\textbf{LaMAR}. %
As shown in \Tab\ref{tab:lamar}, \emph{ImLoc} again achieves the best performance for all scenes and thresholds. The improvement is consistent and substantial across all scenes. 
Note that the CAB scene is a recording of indoor offices and localizers struggle to differentiate similar structure on different floors of the same building, which results in low recall. 
We also evaluate (\Tab\ref{tab:lamar_hololens}, supplementary) \emph{ImLoc} on LaMAR HoloLens data, again achieving the best performance. 

\noindent\textbf{Cambridge Landmarks}. We compare to state-of-the-art feature matching based, and to storage efficient SCR methods in \Tab\ref{tab:results_cam}. 
\emph{ImLoc} achieves state-of-the-art accuracy on all scenes with a map size of 90MB. %
We further explore two more strongly compressed versions (\cf~\ref{sec:ablation_compression}) of \emph{ImLoc}, %
termed \textit{nano} and \textit{micro}, 
with map sizes of only 2MB and 16Mb. 
Notably, our \textit{nano} and \textit{micro} versions outperform SCR methods~\cite{brachmann2023ace,glace2024cvpr} with similar storage arrangements. 
This underlines the remarkable capability of \emph{ImLoc} to trade off accuracy for storage efficiency without losing too much localization performance (also visualized in \Fig\ref{fig:teaser}, \emph{right}). 

\noindent\textbf{Aachen Day-Night}. Following~\cite{panek2025guide}, \Tab\ref{tab:sota_comparison} compares \emph{ImLoc} with state-of-the-art structure-based~\cite{sarlin2019coarse,panek2022meshloc,dong2023lazy} and structureless~\cite{panek2025guide} methods. %
Our method is consistently more accurate on both query at daytime and nighttime.

\subsection{Ablation Study}
\begin{figure}[t]
    \centering
    \includegraphics[trim={0 3 0 28}, clip, width=1.0\linewidth]{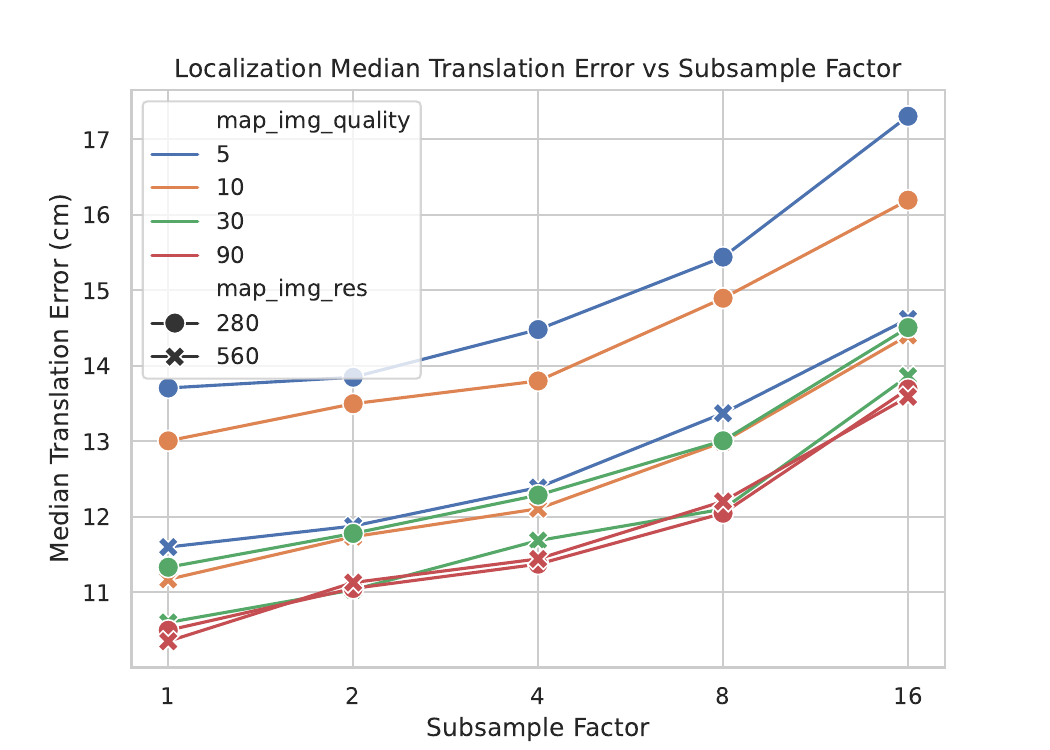}
    \caption{RGB Image Subsampling and Compression on Cambridge. Localization can tolerate low quality setting for modern image compression like JPEG XL but it is more sensitive to keyframe subsampling or downsampling resolution, which decreases performance.
    }
    \label{fig:rgb_compress}
    \vspace{-0.31cm}
\end{figure}
\customparagraph{Compression Potential}
\label{sec:ablation_compression}
We analyze the influence of image, depth and compression level on the map size and accuracy. Our default conservative compression settings are designed to maintain more original information for general use. In this case, each RGB image takes about 60KB, and each depth image about 17KB. However, when exclusively targeting localization, we may compress the map more aggressively. 
We combine the following compression techniques to arrive at the micro version in~\Tab\ref{tab:results_cam}: using an image resolution of $280^2$ JPEG XL~\cite{alakuijala2019jpeg} compressed with quality of 30 and a $70^2$ depth image resolution, quantized to 8bits. We further downsample the number of frames by a factor of 8 to get a nano version.
In this case, \emph{ImLoc} only needs about 2MB for each scene on average, where about 1MB is used for retrieval features stored in half precision without compression, 300 KB for depth, and 700KB for RGB images. For a fair comparison to align with the other baselines, we do not further compress retrieval features. 
Although we observe that the tolerance for compression can be dataset dependent, the trend is exemplary.

\customparagraph{Image compression} \Fig\ref{fig:rgb_compress} shows that using modern image compression like JPEG XL~\cite{alakuijala2019jpeg}, compressing with a low quality setting (30), or 2x downsampling the resolution, does not decrease the localization performance of our method  significantly. 
Further compression with lower quality or combining low quality compression with downsampling decreases the localization performance gradually. However, subsampling keyframes (uniformly by timestamp) will instantly decrease the localization performance. The evaluation is performed on Cambridge Landmarks~\cite{kendall2015posenet}. 
\begin{figure}[t]
    \centering
    \includegraphics[trim={0 5 0 5}, clip, width=1.0\linewidth]{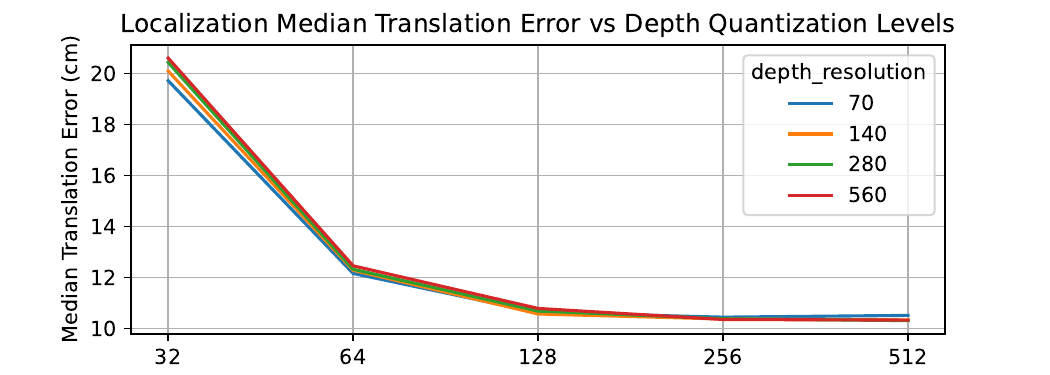}
    \caption{Depth Image Compression on Cambridge. The quantization usually saturates at 8bit quantization (256 depth levels), while a lower depth resolution does not significantly affect performance for any of the selected quantization levels. }
    \label{fig:depth_compress}
    \vspace{-0.35cm}
\end{figure}

\vspace{-0.35cm}
\customparagraph{Depth Resolution}
\Fig\ref{fig:depth_compress} evaluates the effect of choosing resolution and quantization level for the depth images by comparing the median translation error on the Cambridge dataset. 
Although we observe that the tolerance for compression can be data set dependent, the trend is exemplary. 
Higher quantization than 8 bits (256 depth levels) is unnecessary. 
For our default resolution ranging from $560^2$ to $280^2$ pixels we do not observe a statistically significant drop in the accuracy. 
On the contrary, \Fig\ref{fig:depth_compress} shows that we can effectively trade-off memory efficiency for a small drop in accuracy. 
This is further confirmed in \Fig\ref{fig:teaser}, where we can adjust the compression to a chosen memory footprint and
\emph{ImLoc} shows the best performance at the desired map size.

\customparagraph{Depth Quantization}
8 bit quantization for 0.25m to 128m depth has less than 1.4\% relative quantization error, which is sufficient for localization (\Fig\ref{fig:depth_compress}). Especially if map images have similar viewing directions, depth can be quantized vigorously.

\begin{figure}[t]
    \centering
    \includegraphics[trim={0 7 0 7}, clip, width=1.0\linewidth]{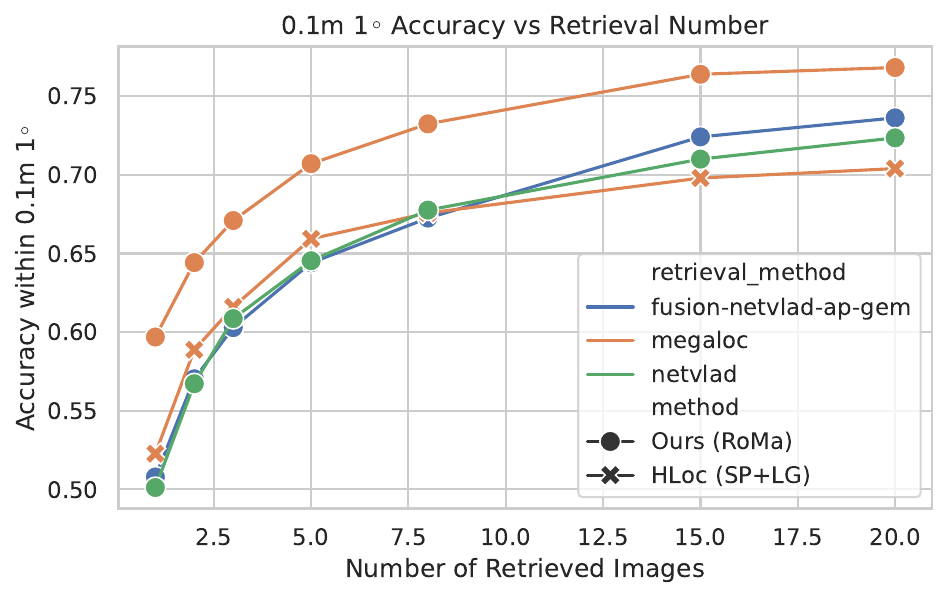}
    \caption{Performance of different retrieval methods and number of retrieved images on LaMAR~\cite{sarlin2022lamar}. We plot the percentage of poses with error smaller than 0.1m and 1$^{\circ}$. }
    \label{fig:depth_compress2}
    \vspace{-0.36cm}
\end{figure}

\customparagraph{Image Retrieval}
\begin{figure}[t]
    \centering
    \includegraphics[trim={0 7 0 7}, clip, width=1.0\linewidth]{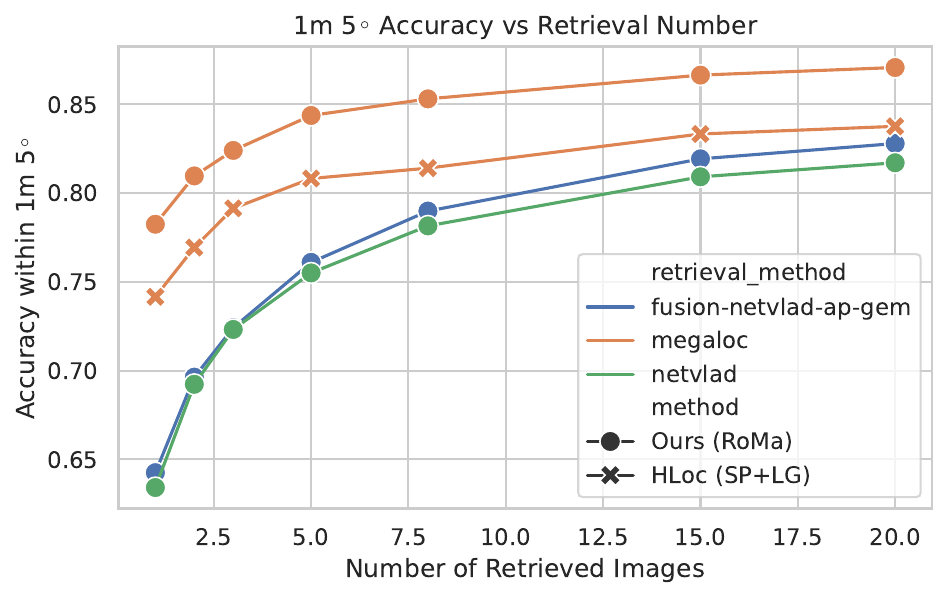}
    \caption{Performance of different retrieval methods and number of retrieved images on LaMAR~\cite{sarlin2022lamar}. We plot the percentage of poses with error smaller than 1m and 5$^{\circ}$.
    }
    \label{fig:depth_compress1}
    \vspace{-0.35cm}
\end{figure}
\begin{table}[b]
    \begin{center}
        \resizebox{\linewidth}{!}{
            \begin{tabular}{lccccccccccccccccccccc}
                \toprule
                \multicolumn{1}{l}{\multirow{2}{*}{Geometry}} &
                \multicolumn{2}{c}{CAB (val)}                &                &
                \multicolumn{2}{c}{HGE (val)}                &                &
                \multicolumn{2}{c}{LIN (val)}                &                &

                \\

                \cline{2-3} \cline{5-6} \cline{8-9} \cline{11-12} 
                                                             & (1, 0.1)       & (5, 1.0)       &  & (1, 0.1)       & (5, 1.0)       &  & (1, 0.1)       & (5, 1.0)       \\
                \midrule
                Sparse                         & 52.78          & 61.62          &  & 69.54          & 94.33          &  & 88.76          & 95.37          \\
                Dense                   & \textbf{58.84} & \textbf{66.16} &  & \textbf{73.73} & \textbf{95.38} &  & \textbf{90.08} & \textbf{95.70} \\
                \bottomrule
            \end{tabular}
        }
    \end{center}
    \vspace{-0.3cm}
    \caption{By not committing to a sparse set of points we can improve the localization accuracy via matching on a denser set of points when compared to a sparse set.}
    \label{tab:abla_geometry}
    \vspace{-0.25cm}
\end{table}
\Fig\ref{fig:depth_compress1} and \Fig\ref{fig:depth_compress2} analyze the performance of different retrieval
methods and the effect of choosing the number of retrieved images on the LaMAR dataset.
Shown are the percentage of poses with an error smaller than 0.1m and 1$^{\circ}$ (\Fig\ref{fig:depth_compress1}) and 1m and 5$^{\circ}$ (\Fig\ref{fig:depth_compress2}).
On challenging datasets like LaMAR, a good image retrieval model is important. We observe that MegaLoc leads to significant improvements compared to previous retrieval methods. 
In general, the accuracy for both thresholds, coarse and fine, increases as more images are retrieved. 
When given the same (number of) retrieved images our method can utilize the information of each retrieved image more effectively, 
compared to the currently best sparse method (HLoc(SP+LG))~\cite{sarlin2019coarse,detone2018superpoint,lindenberger2023lightglue}. 
We can achieve a higher final accuracy when a sufficient number of images are retrieved. 

\customparagraph{Comparison of different geometric representations}
Tab~\ref{tab:abla_geometry} validates the use of dense matching at test time by matching only at the sparse pixels selected by superpoint~\cite{detone2018superpoint}. 
We compare this to matching all the points. 
Our dense depth representation allows us to fully utilize the power of dense matchers. 
By not sparsifying potential matches prematurely and postponing the decision which points to use as correspondences until after matching, we can obtain a higher accuracy. 

\subsection{Runtime}
The overall runtime can be broken down into the following components: image pair matching, pose computation, image retrieval and nearest neighbours search. 
By accelerating the inference with training free optimization (see supplementary), dense RoMa matching of $560\!\times\!560$ images (we upsample if stored at lower resolution) takes about 80ms per pair. The timings for solving for the pose, depend on the actual number of correspondences and the inlier ratio. 
Our RANSAC loop usually takes about 200ms. 
The time for extracting retrieval features depends on the model. It usually takes about 50ms for Megaloc~\cite{berton2025megaloc}. The NN Search of retrieval features can be done very fast on the GPU and usually takes less than 1ms. 
All the reported numbers are obtained on an NVIDIA 4090 GPU.

%% file: table/lamar.tex
\begin{table*}
    \begin{center}
        \resizebox{\textwidth}{!}{
            \begin{tabular}{lccccccccccccccccccccccc}
                \toprule
                \multicolumn{1}{l}{\multirow{2}{*}{Methods}} &
                \multicolumn{2}{c}{CAB (val)}                &                &
                \multicolumn{2}{c}{HGE (val)}                &                &
                \multicolumn{2}{c}{LIN (val)}                &                &
                \multicolumn{2}{c}{Avg (val)}                &                &
                \multicolumn{2}{c}{CAB (test)}               &                &
                \multicolumn{2}{c}{HGE (test)}               &                &
                \multicolumn{2}{c}{LIN (test)}               &                &
                \multicolumn{2}{c}{Avg (test)}                                                                                                                                                                                                                                                                                                                    \\

                \cline{2-3} \cline{5-6} \cline{8-9} \cline{11-12} \cline{14-15} \cline{17-18} \cline{20-21} \cline{23-24}
                                                             & (1, 0.1)       & (5, 1.0)       &  & (1, 0.1)       & (5, 1.0)       &  & (1, 0.1)       & (5, 1.0)       &  & (1, 0.1)       & (5, 1.0)       &  & (1, 0.1)       & (5, 1.0)       &  & (1, 0.1)       & (5, 1.0)       &  & (1, 0.1)       & (5, 1.0)       &  & (1, 0.1)       & (5, 1.0)       \\
                \midrule
                Hloc (SP+SG)                                 & 52.02          & 60.61          &  & 70.59          & 94.12          &  & 83.80          & 94.88          &  & 68.80          & 83.20          &  & 56.90          & 71.90          &  & 55.60          & 73.50          &  & 76.50          & 90.20          &  & 63.00          & 78.53          \\
                Hloc (SP+LG)                                 & 49.75          & 58.08          &  & 71.01          & 92.44          &  & 84.30          & 94.88          &  & 68.35          & 81.80          &  & 54.60          & 70.20          &  & 56.30          & 72.70          &  & 76.10          & 89.20          &  & 62.33          & 77.37          \\
                \midrule
                Ours (RoMa)                                  & \textbf{58.84} & \textbf{66.41} &  & \textbf{73.95} & \textbf{95.38} &  & \textbf{89.92} & \textbf{95.87} &  & \textbf{74.24} & \textbf{85.89} &  & \textbf{63.80} & \textbf{79.50} &  & \textbf{60.90} & \textbf{80.20} &  & \textbf{80.10} & \textbf{92.70} &  & \textbf{68.27} & \textbf{84.13} \\
                \bottomrule
            \end{tabular}
        }
    \end{center}
    \vspace{-0.3cm}
    \caption{\textbf{Results on LaMAR dataset}, computed on each of the three scenes, for Phone queries on validation set and submitted on the benchmark to obtain test set results. For each scene, we report the recall at (1°, 0.1m) and (5°, 1.0m), following the LaMAR paper~\cite{sarlin2022lamar}. We use 50 top-retrieved images for mapping and 10 top-retrieved images for localization using Megaloc~\cite{berton2025megaloc}. }
    \label{tab:lamar}
\end{table*}

%% file: table/cambridge.tex
\begin{table*}[]
  \centering
  \footnotesize
  \begin{tabular}{clccccccc}
    \toprule
    \multicolumn{1}{l}{} &                                                                                                                                                                         &                                                                      & \multicolumn{5}{c}{Cambridge Landmarks}                                                                                          \\
    \cmidrule(l){4-8}
    \multicolumn{1}{l}{} &                                                                                      &  \multirow{-2}{*}{\begin{tabular}[c]{@{}c@{}}Map\\ Size\end{tabular}} & Court                                   & King's          & Hospital        & Shop            & St. Mary's
                         & \multirow{-2}{*}{\begin{tabular}[c]{@{}c@{}}Average \\ (cm / $^\circ$)\end{tabular}}                                                                                                                                                                                                                                                                                               \\
    \midrule
    \multirow{7}{*}{\rotatebox{90}{FM}}
                         & AS (SIFT) \cite{Sattler2017AS}                                                                                                                                       & $\sim$200MB                                                          & 24/0.1                                  & 13/0.2          & 20/0.4          & \textbf{4/0.2}  & 8/0.3          & 14/0.2          \\
                         & Hloc (SP+SG) \cite{detone2018superpoint, sarlin2020superglue}                                                                                                        & $\sim$800MB                                                          & \textbf{16/0.1}                         & {\ul 12/0.2}    & {\ul15/0.3}     & \textbf{4/0.2 } & \textbf{7/0.2} & {\ul 11/0.2}    \\
                         & Hloc (SP+LG)~\cite{detone2018superpoint,lindenberger2023lightglue}                                                                                                   & $\sim$800MB                                                          & 18/0.1                                  & 12/0.2          & {\ul15/0.3}     & \textbf{4/0.2}  & \textbf{7/0.2} & {\ul11/0.2}     \\
                         & Hloc (Aliked+LG)~\cite{zhao2023aliked, lindenberger2023lightglue}   & $\sim$660MB                                                                                                                                                                        & 19/0.1                                  & 13/0.2          & 16/0.3          & \textbf{4/0.2}  & \textbf{7/0.2} & 12/0.2          \\
                         & Hloc (RoMa)~\cite{edstedt2023roma}                                                                                                                                   &                                               $\sim$300MB                       & {\ul 17/0.1}                            & 15/0.2          & 17/0.3          & 7/0.3           & {\ul 8/0.2}    & 13/0.2          \\
                         & pixLoc \cite{sarlin21pixloc}                                                                                                                                         & $\sim$600MB                                                          & 30/0.1                                  & 14/0.2          & 16/0.3          & {\ul5/0.2}      & 10/0.3         & 15/0.2          \\
                         & GoMatch \cite{zhou2022gomatch}                                                                                                                                       & $\sim$12MB                                                           & N/A                                     & 25/0.6          & 283/8.1         & 48/4.8          & 335/9.9        & N/A             \\
                         & HybridSC \cite{compression2019cvpr}                                                                                                                                  & \textbf{$\sim$1MB}                                                            & N/A                                     & 81/0.6          & 75/1.0          & 19/0.5          & 50/0.5         & N/A             \\
    \midrule
    \multirow{6}{*}{\rotatebox{90}{SCR}}
                         & DSAC* (Full) \cite{Brachmann2021dsacstar}                                                                                                                         & 28MB                                                                 & 49/0.3                                  & 15/0.3          & 21/0.4          & 5/0.3           & 13/0.4         & 21/0.3          \\
                         & SANet \cite{yang2019sanet}                                                                                                                                        & $\sim$260MB                                                          & 328/2.0                                 & 32/0.5          & 32/0.5          & 10/0.5          & 16/0.6         & 84/0.8          \\
                         & SRC \cite{dong2022visual}                                                                                                                                         & 40MB                                                                 & 81/0.5                                  & 39/0.7          & 38/0.5          & 19/1.0          & 31/1.0         & 42/0.7          \\
                         & ACE~\cite{brachmann2023ace}                                                                                                                                          & 4MB                                                                  & {43/0.2}                                & 28/0.4          & {31/0.6}        & 5/0.3           & {18/0.6}       & 25/0.4          \\
                         & Poker (ACE~\cite{brachmann2023ace} $\times$ 4)                                                                                                                       & 16MB                                                                 & 28/0.1                                  & 18/0.3          & 25/0.5          & 5/0.3           & 9/0.3          & 17/0.3          \\
                         & GLACE~\cite{glace2024cvpr}                                                                                                                                           & 13MB                                                                 & 19/0.1                                  & 19/0.3          & 17/0.4          & \textbf{4/0.2}  & 9/0.3          & 14/0.3          \\
    \midrule
                         & Ours                                                                                                                                                                 & $\sim$90MB                                                           & \textbf{16/0.1}                         & \textbf{11/0.2} & \textbf{14/0.3} & \textbf{4/0.2}  & \textbf{7/0.2} & \textbf{10/0.2} \\
                         & Ours (Micro)                                                                                                                                                          & $\sim$16MB                                                            & \textbf{16/0.1}                                  & 12/0.2          & 16/0.3          & \textbf{4/0.2}           & 8/0.3          & 11/0.2          \\
                         & Ours (Nano)                                                                                                                                                          & {\ul$\sim$2MB}                                                            & 18/0.1                                  & 15/0.3          & 19/0.4          & 5/0.3           & 9/0.3          & 13/0.3          \\
    \bottomrule
  \end{tabular}
  \caption{\textbf{Results on Cambridge Landmarks \cite{kendall2015posenet}.} We report median rotation and position errors. Best results are in \textbf{bold}, second-best results are \underline{underlined}. For image retrieval, we use 10 images retrieved by NetVLAD~\cite{arandjelovic16netvlad} for our method.}
  \label{tab:results_cam}
  \vspace{-0.205cm}
\end{table*}

%% file: table/aachenv11.tex
\begin{table}[b!]
\vspace{-0.155cm}
    \centering
    \small
    \resizebox{\linewidth}{!}{
    \begin{tabular}{l ll l l}
        \hline
        Methods                                       & Matcher            & Day                         & Night                       \\ \hline
        \multirow{2}{*}{Hloc~\cite{sarlin2019coarse}} & SP+SG               & 88.1 / 95.4 / 98.9          & 73.3 / 87.4 / 97.9          \\
                                                      & SP+LG               & 87.0 / 94.8 / 98.5          & 70.2 / 87.4 / 97.4          \\ \hline
        MeshLoc~\cite{panek2022meshloc}              & SP+LG                 & 84.2 / 92.5 / 98.5          & 70.2 / 85.9 / 96.9          \\
        \hline
        LazyLoc~\cite{dong2023lazy}                   & SP+LG                    & 76.8 / 87.7 / 94.7          & 58.1 / 84.3 / 94.2          \\ \hline

        \multirow{2}{*}{E5+1~\cite{panek2025guide}}                         & SP+LG    & 76.6 / 88.3 / 97.5          & 61.3 / 85.9 / 96.9          \\
                                                      & RoMa     & 78.4 / 89.8 / 97.8          & 65.4 / 84.8 / 97.9          \\
        \hline
        Local tri.~\cite{panek2025guide}                           & SP+LG                   & 83.5 / 91.4 / 97.8          & 66.5 / 84.3 / 96.3          \\
        Local tri.~\cite{panek2025guide}                           & RoMa                    & 84.0 / 92.8 / 97.9          & 68.6 / 85.3 / 97.9          \\
        \hline
        Ours                                          & RoMa                    & \textbf{89.3 / 96.1 / 99.3} & \textbf{74.3 / 91.6 / 99.0} \\
        \hline
    \end{tabular}
    }
    \caption{\textbf{Results on on Aachen Day-Night v1.1~\cite{sattler2018cvpr, sattler2012bmvc}}. We compare state-of-the-art structure-based and structureless~\cite{panek2025guide} approaches. EigenPlaces~\cite{berton2023eigenplaces} is used to retrieve the top-10 images and we report localization recall at thresholds of (0.25m, 2°) / (0.5m, 5°) / (5m, 10°). %
    }
    \label{tab:sota_comparison}
\end{table}

%% file: sec/5_conclusion.tex
\section{Conclusion}
\label{sec:conclusion}

We have revisited visual localization through the lens of a simple, yet powerful image-based representation, augmenting posed RGB images with precomputed depth maps. 
By leveraging dense depth maps, we enable effective geometric reasoning and robust pose estimation, while maintaining a compact and easily updatable scene representation. 
Extensive experiments on various challenging benchmarks demonstrate that our pipeline achieves state-of-the-art accuracy across diverse conditions, outperforming %
more complex structure-based and NVS approaches, while offering advantages in storage efficiency and adaptability. 
We believe that this work opens new directions for scalable, efficient, and robust localization systems, 
and provides a strong baseline for future research in visual localization.

%% file: sec/X_suppl.tex
\clearpage
\setcounter{page}{1}
\maketitlesupplementary
\section{Accelerating the RoMa Matcher}
Our baseline is RoMa~\cite{edstedt2023roma} in its default setting, using its custom CUDA kernel for the local correlation operation. 
Running the bidirectional matching (\ie, query-to-map and map-to-query) with RoMa at $560^2$ resolution takes 126ms per pair on an NVIDIA RTX 4090 GPU. 
We show that it can achieve a $1.77\times$ speedup to 71ms per pair with training-free inference acceleration.

\customparagraph{PyTorch JIT Compilation}
We first leverage PyTorch's JIT compilation to accelerate the feature extraction and matching modules of RoMa during inference, achieving a run time of 102ms per pair.

\customparagraph{Batch Processing}
When matching query images with multiple retrieved mapping images, we can process all image pairs in a single batch to better utilize the parallel processing capabilities of GPUs. Specifically, we use a maximum batch size of 20 pairs and perform bidirectional matching for each pair. 
It takes 1.7s in total for a batch with 20 pairs,~\ie, 85ms per pair.

\customparagraph{Feature Extraction}
When matching a query image with multiple retrieved images, we only need to extract the feature of the query image once, and reuse it for all the matching pairs. For a 20-pair batch, this results in 1.58s, or 79ms per pair.

\customparagraph{Convolution Padding}
Finally, we observe that some layers in the RoMa refinement module use convolutions with channel size not divisible by 8. This hinders  efficient utilization of Tensor Cores on NVIDIA GPUs. 
Without retraining, we round up the channel size to the nearest multiple of 8 by padding zeros to the convolution weights and input feature maps. With this modification, a 20-pair batch consumes 1.42s, or 71ms per image pair.

\section{Additional Compression Statistics}
\begin{figure}[tb]
    \centering
    \includegraphics[trim={0 0 0 40}, clip, width=1.0\linewidth]{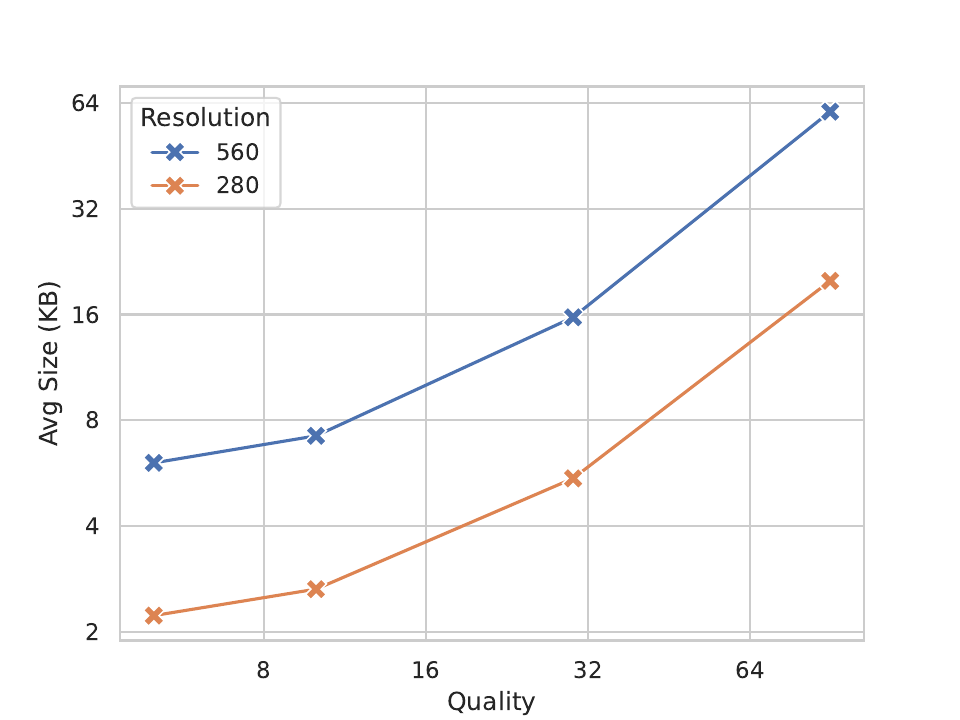}
    \caption{\textbf{RGB Image Compression quality versus image size on Cambridge Landmarks~\cite{kendall2015posenet}}. Note that 1000 128-dimensional local descriptors (\cite{detone2018superpoint,zhao2023aliked,tyszkiewicz2020disk}) require 256KB, while a 4096-dimensional global descriptor (NetVLAD~\cite{arandjelovic16netvlad}) takes 8KB,  when stored in half-precision. With modern image compression (JPEG XL~\cite{alakuijala2019jpeg}), storing the RGB image itself is smaller than storing local descriptors and even global descriptors.}
    \label{fig:rgb_jxl_sizestat}
    \vspace{-0.15cm}
\end{figure}
\begin{figure}[t]
    \centering
    \includegraphics[width=1.0\linewidth]{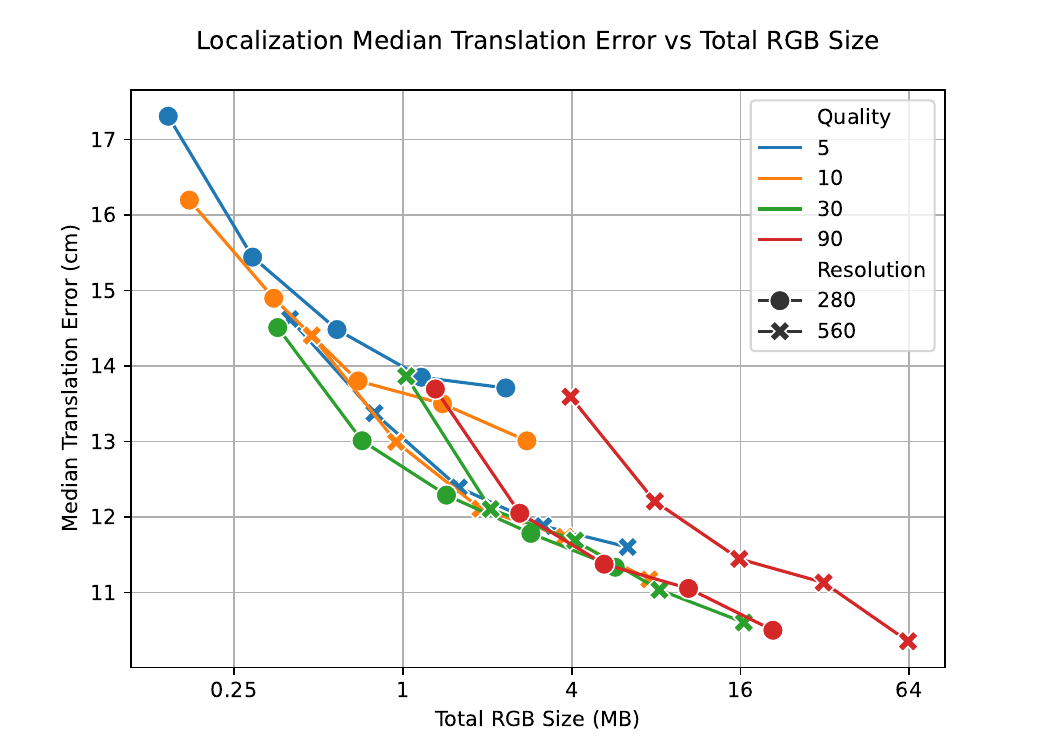}
    \caption{\textbf{RGB Image Compression techniques versus pose accuracy on Cambridge Landmarks~\cite{kendall2015posenet}}. The markers ($\bullet$ and $\times$) in each line depict keyframe subsampling factors $k$ of 16, 8, 4, 2, 1. Subsampling is performed by sorting the images by their filename and selecting one image every $k$ images. 
    By adjusting image compression quality, image resolution, and subsampling keyframes, we can conveniently trade off between storage and localization accuracy.}
    \label{fig:rgb_jxl_size_vs_err}
    \vspace{-0.15cm}
\end{figure}
In \Fig\ref{fig:rgb_jxl_sizestat}, we plot the average image size against the compression quality for resolutions of $560^2$ and $280^2$. 
We remark that storing 1000 128-dimensional local descriptors (\cite{detone2018superpoint,zhao2023aliked,tyszkiewicz2020disk}) requires 256KB per image, while storing a 4096-dimensional global descriptor (NetVLAD~\cite{arandjelovic16netvlad}) takes 8KB when stored in half-precision. 
Hence, storing high-dimensional local descriptors is significantly more expensive than storing the entire image with compression (JPEG XL~\cite{alakuijala2019jpeg}).
Evaluating compression techniques for sparse descriptors or exploring more efficient descriptors for the task of image-based localization is beyond the scope of this paper.

\begin{figure}[t!]
    \centering
    \includegraphics[width=1.0\linewidth]{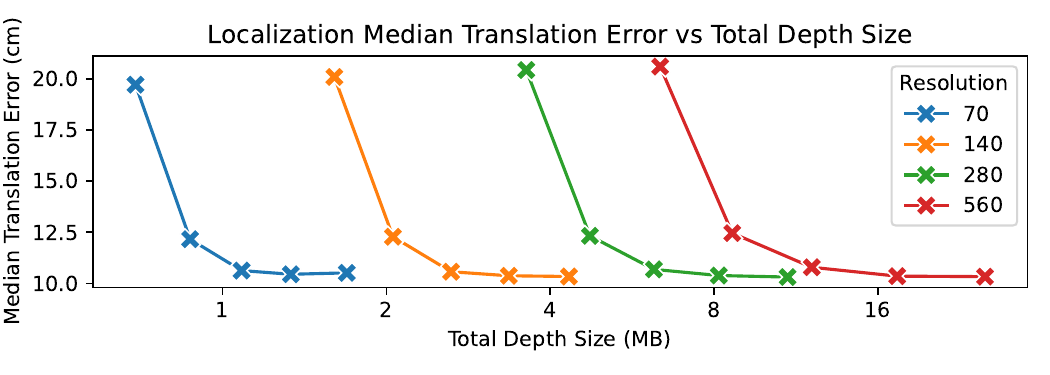}
    \caption{\textbf{Depth Image Compression techniques versus pose accuracy on Cambridge Landmarks~\cite{kendall2015posenet}}. The markers ($\times$) in each line depict quantization into 32, 64, 128, 256, 512 levels (5 to 9 bits).}
    \label{fig:depth_jxl_size_vs_err}
    \vspace{-0.15cm}
\end{figure}

We compare different image compression methods \wrt their median translation error for RGB images in \Fig\ref{fig:rgb_jxl_size_vs_err} and for depth images in \Fig\ref{fig:depth_jxl_size_vs_err} on Cambridge Landmarks. 

For RGB images, we consider keyframe subsampling factors between 16,8,4,2 and 1, image sizes of $560^2$ and $280^2$, and compression qualities of 5,10,30, and 90. 
Keyframe subsampling is performed by sorting images by filename and selecting every $k^\textrm{th}$ frame. 
In \Fig\ref{fig:rgb_jxl_size_vs_err}, we observe that images usually do not require both high quality and high resolution; a combination of moderate compression techniques is often preferable. 
Moving along the Pareto frontier allows us to find optimal combinations for different targeted memory budgets. 

For depth images, \Fig\ref{fig:depth_jxl_size_vs_err} shows the effect of various image resolutions and quantization levels of 32, 64, 128, 256, and 512 (5 to 9 bits). 
We observe that reducing the image resolution should be preferred over all other measures when aiming to minimize the memory consumption of the map. 
While quantization levels above 8 bits show little benefit, using 7 instead of 8 bits only slightly reduces localization accuracy. Using fewer bits leads to noticeable performance degradation. 

\begin{figure}[t!]
    \centering
    \includegraphics[trim={0 0 0 15}, clip,width=1.0\linewidth]{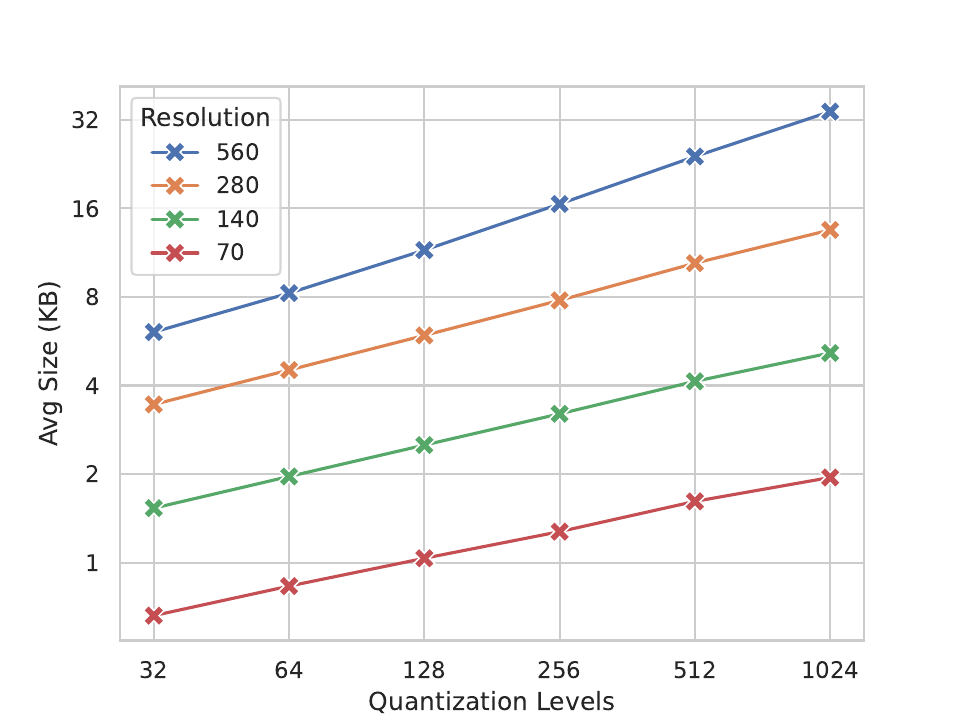}
    \caption{\textbf{Depth Image Compression quality versus image size on Cambridge Landmarks~\cite{kendall2015posenet}}. Depth with suitable quantization (default: 8bit) can be compressed well, even using lossless compression. The size is usually similar to low-quality RGB images at the same resolution, and much smaller than a high-quality RGB image. Storing depth along RGB is not a memory bottleneck for \emph{ImLoc}.
    }
    \label{fig:depth_jxl_sizestat}
\end{figure}

Finally, \Fig\ref{fig:depth_jxl_sizestat} illustrates the average depth-image memory consumption for different quantization levels and depth-image resolutions. 
Our quantized depth images compress well under lossless compression. Their sizes are similar to RGB images at the same resolution but compressed (lossy) with low quality, and substantially smaller than RGB images compressed with higher quality. We conclude that storing dense depth maps is not a memory bottleneck for \emph{ImLoc}.

\section{LaMAR Hololens Results}
We also implement a GPU-accelerated generalized absolute pose estimator following poselib~\cite{poselib} for the evaluation of hololens queries with multi-camera rig setup on LaMAR~\cite{sarlin2022lamar} dataset. As shown in \Tab\ref{tab:lamar_hololens}, our method again achieves the best performance for all scenes and thresholds.
\input{table/lamar_hololens.tex}

\section{Flexible Query Matching}
Our dense representation does not only fully utilize the power of dense matcher, but also provides the flexibility to switch to semi-dense or sparse matchers if desired. 
In \Tab\ref{tab:cam_matcher_ablation}, we show the results of running different feature matchers to establish query-map correspondences. While RoMa~\cite{edstedt2023roma} performs the best, semi-dense matching (LoFTR~\cite{sun2021loftr}) and sparse matching (SP~\cite{detone2018superpoint}+LG~\cite{lindenberger2023lightglue}) also deliver good results. 
By comparing \emph{ImLoc} and HLoc equipped with the same matchers, we observe that \emph{ImLoc} consistently outperforms HLoc on the benchmark, while maintaining a lower memory footprint for the map.

\section{Visualization of RoMa Confidence}
\begin{figure*}[t]
    \centering
        \includegraphics[width=0.23\linewidth]{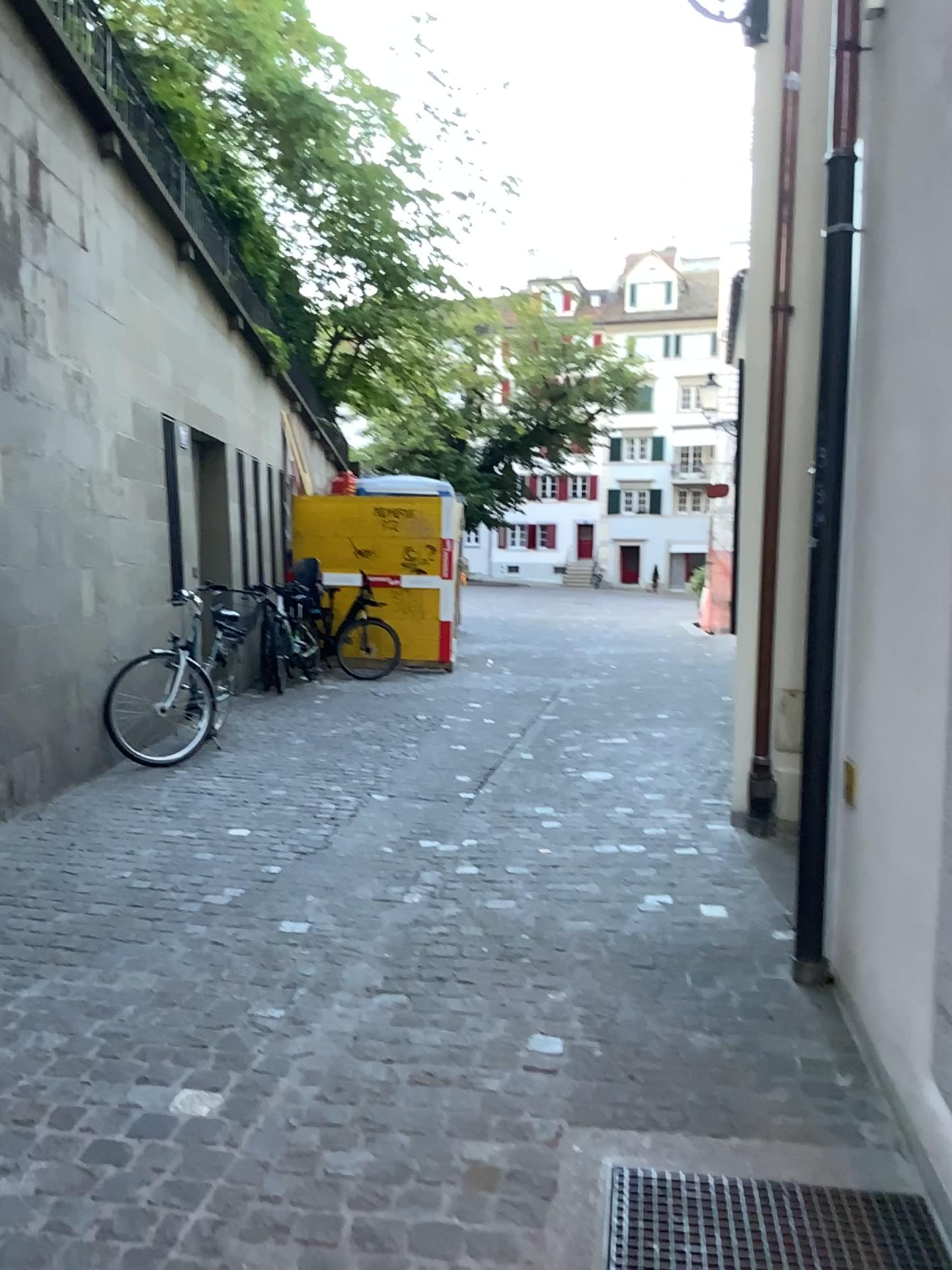}
    \includegraphics[width=0.23\linewidth]{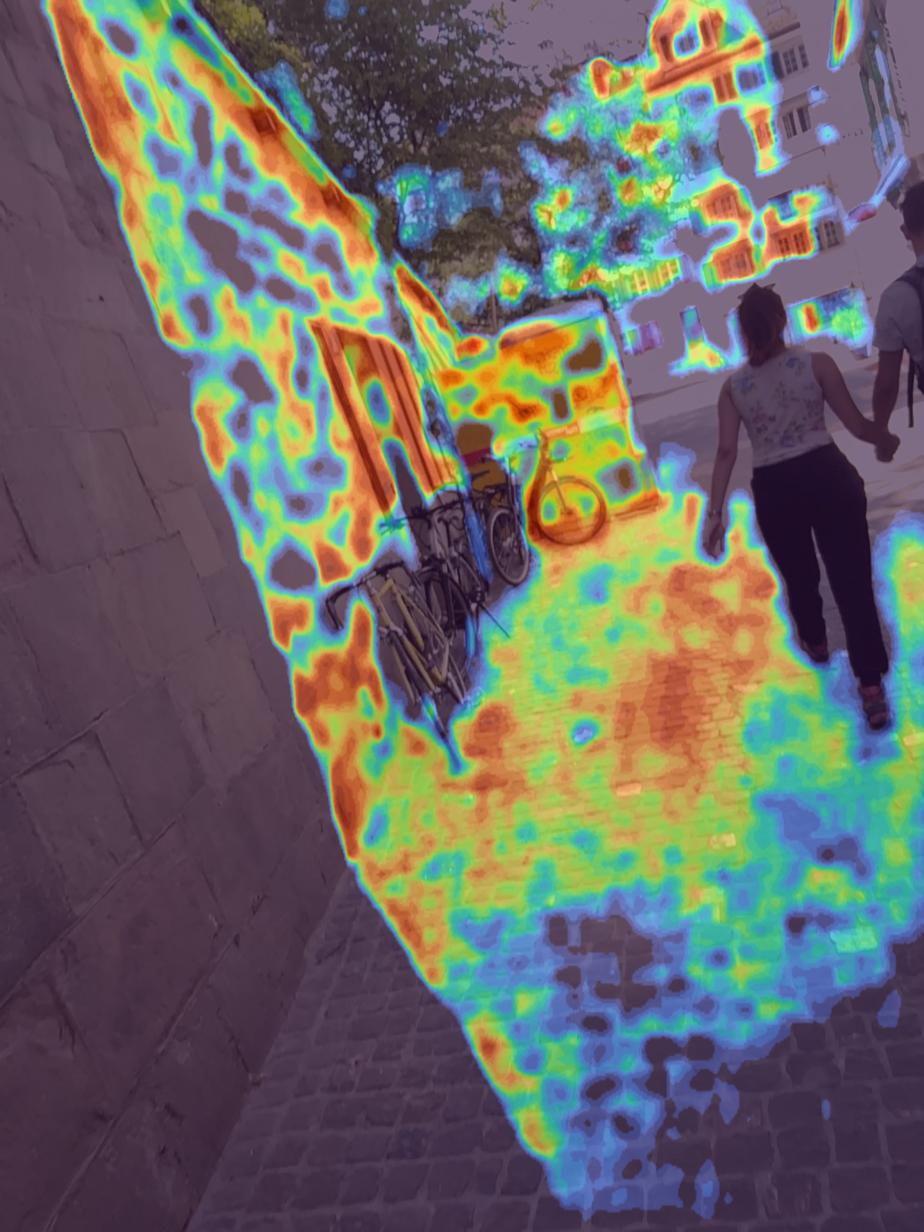}
    \includegraphics[width=0.23\linewidth]{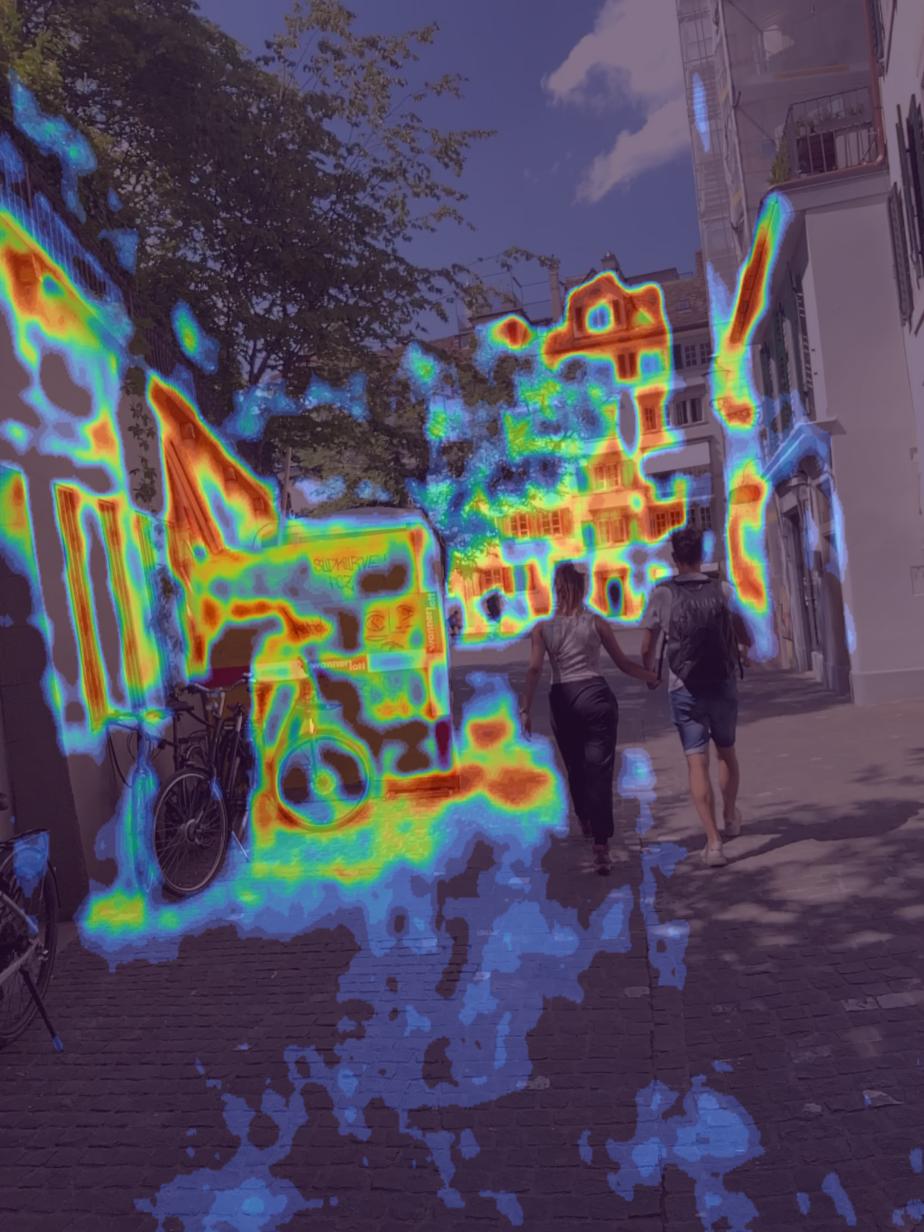}
    \includegraphics[width=0.23\linewidth]{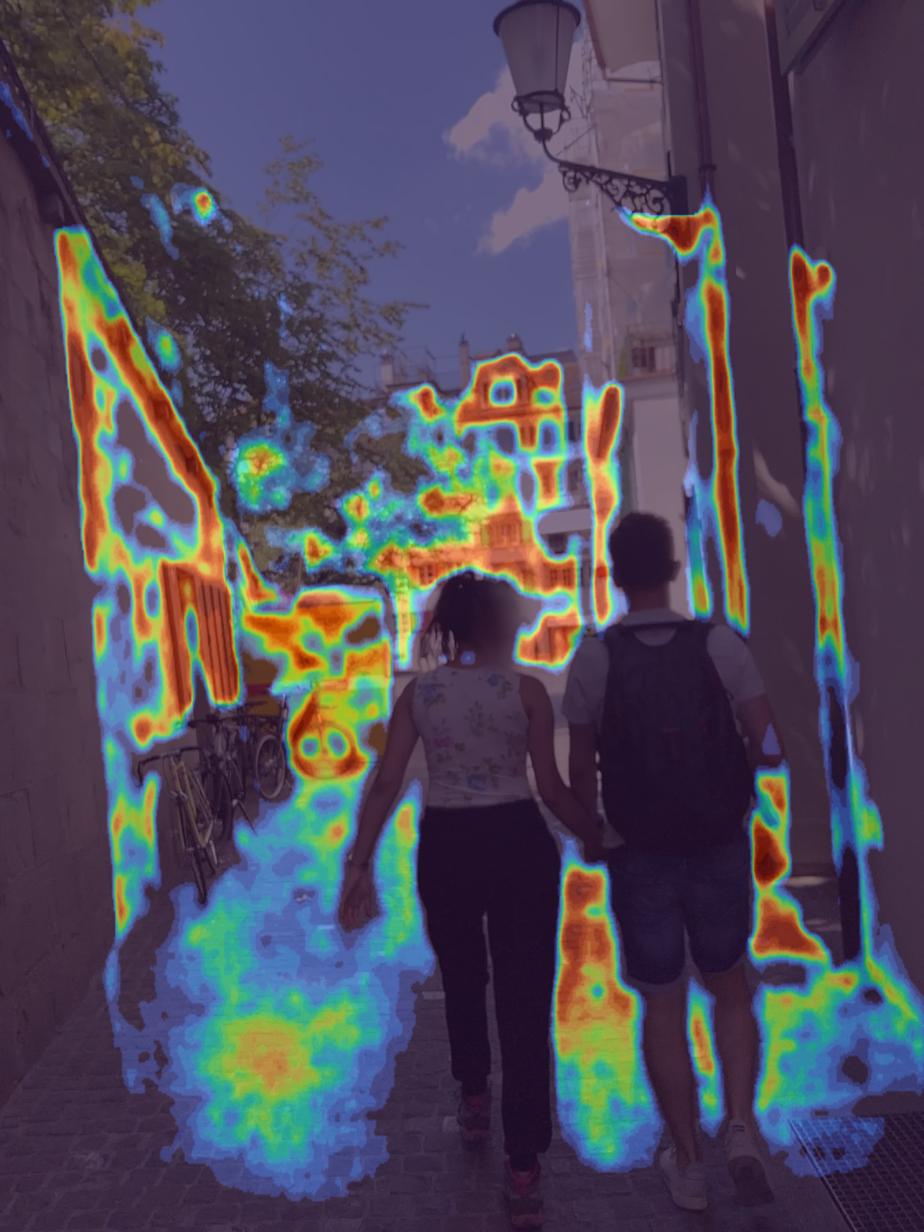}
    \includegraphics[width=0.23\linewidth]{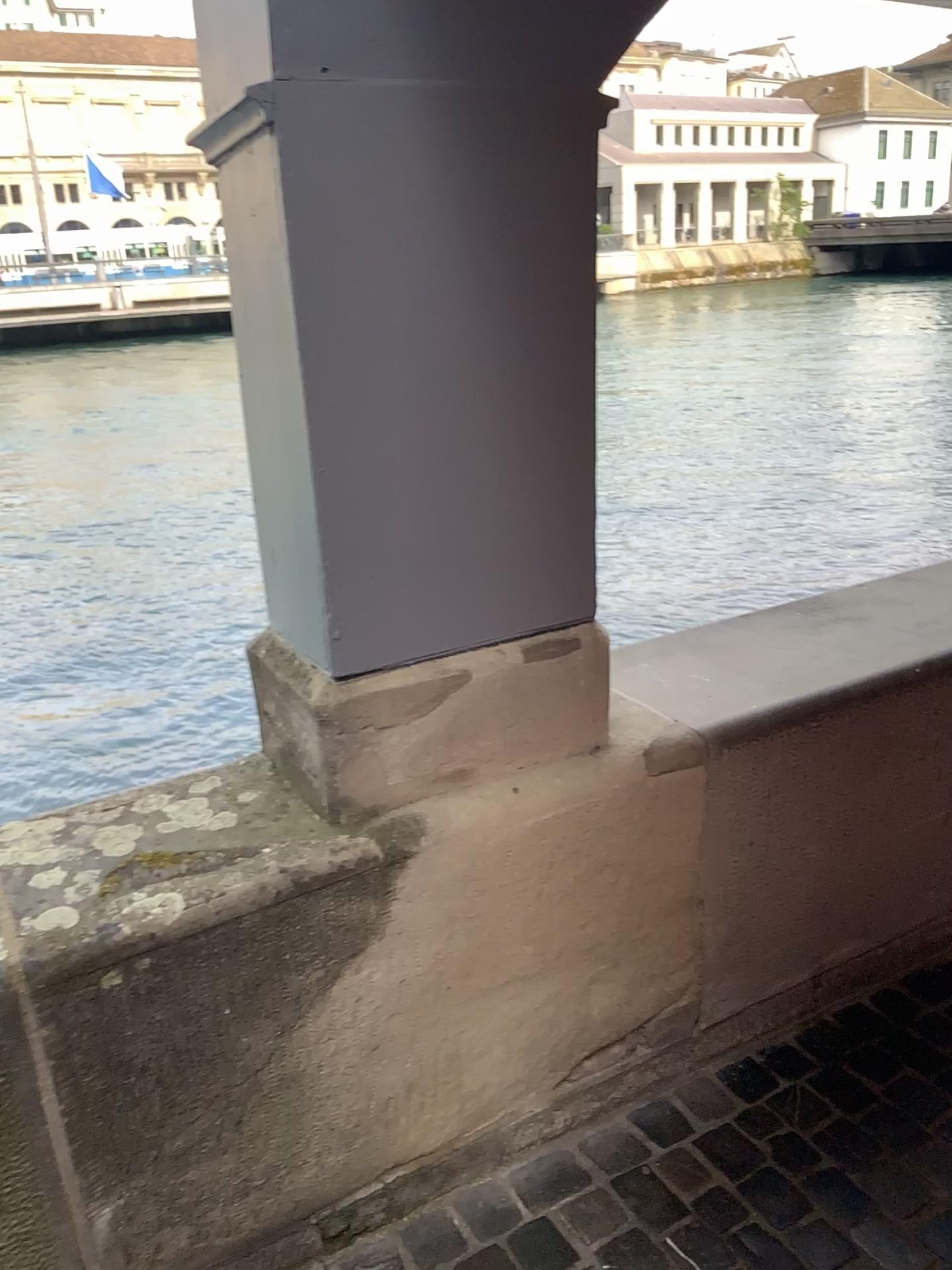} \includegraphics[width=0.23\linewidth]{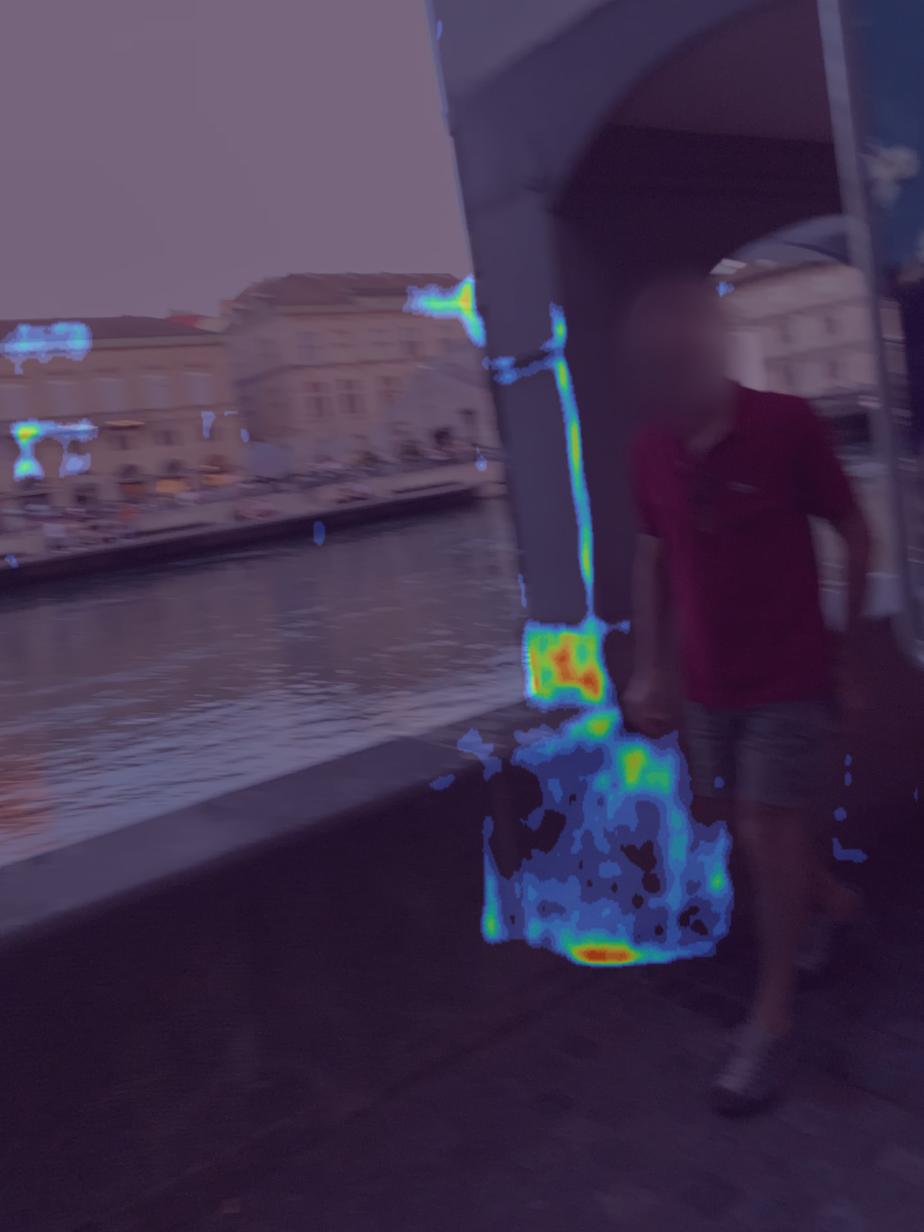}
    \includegraphics[width=0.23\linewidth]{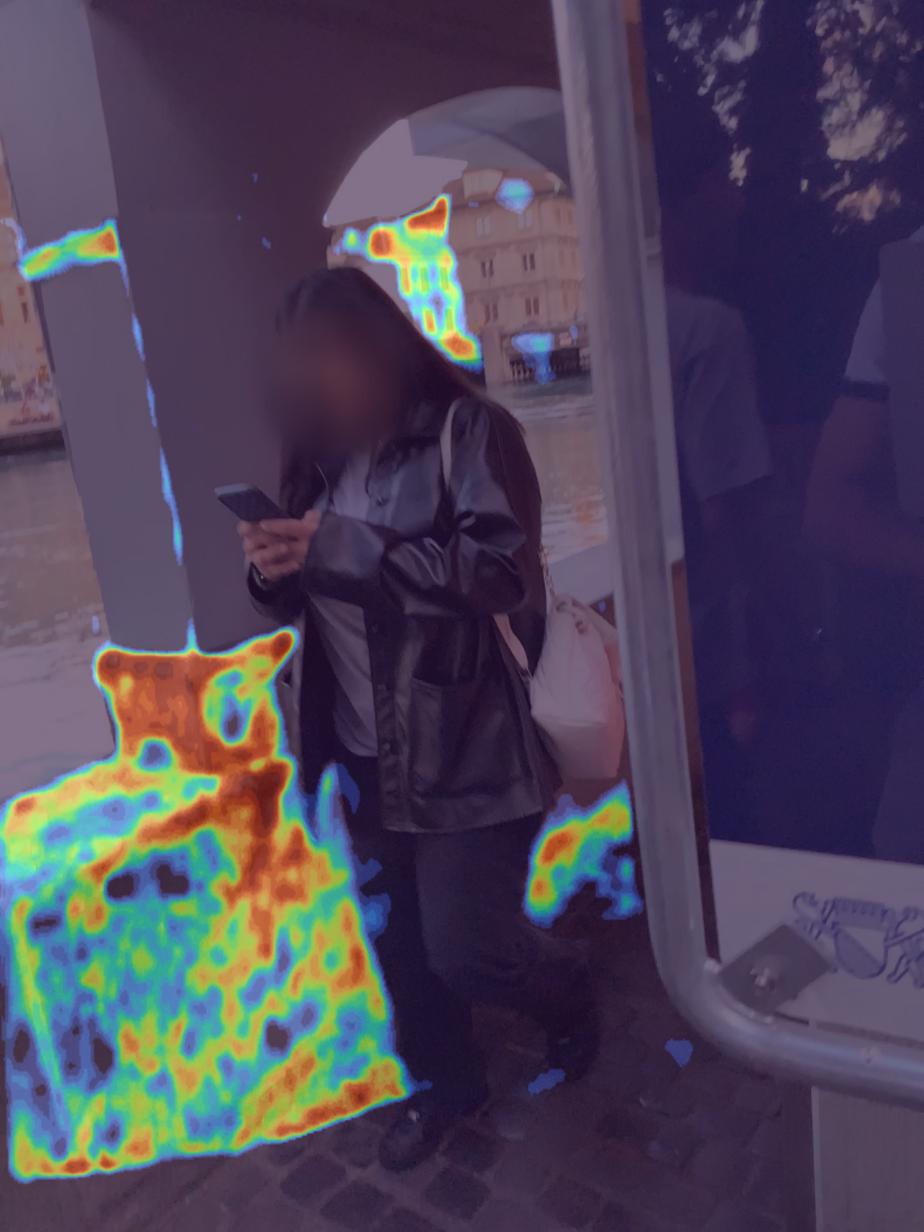}
    \includegraphics[width=0.23\linewidth]{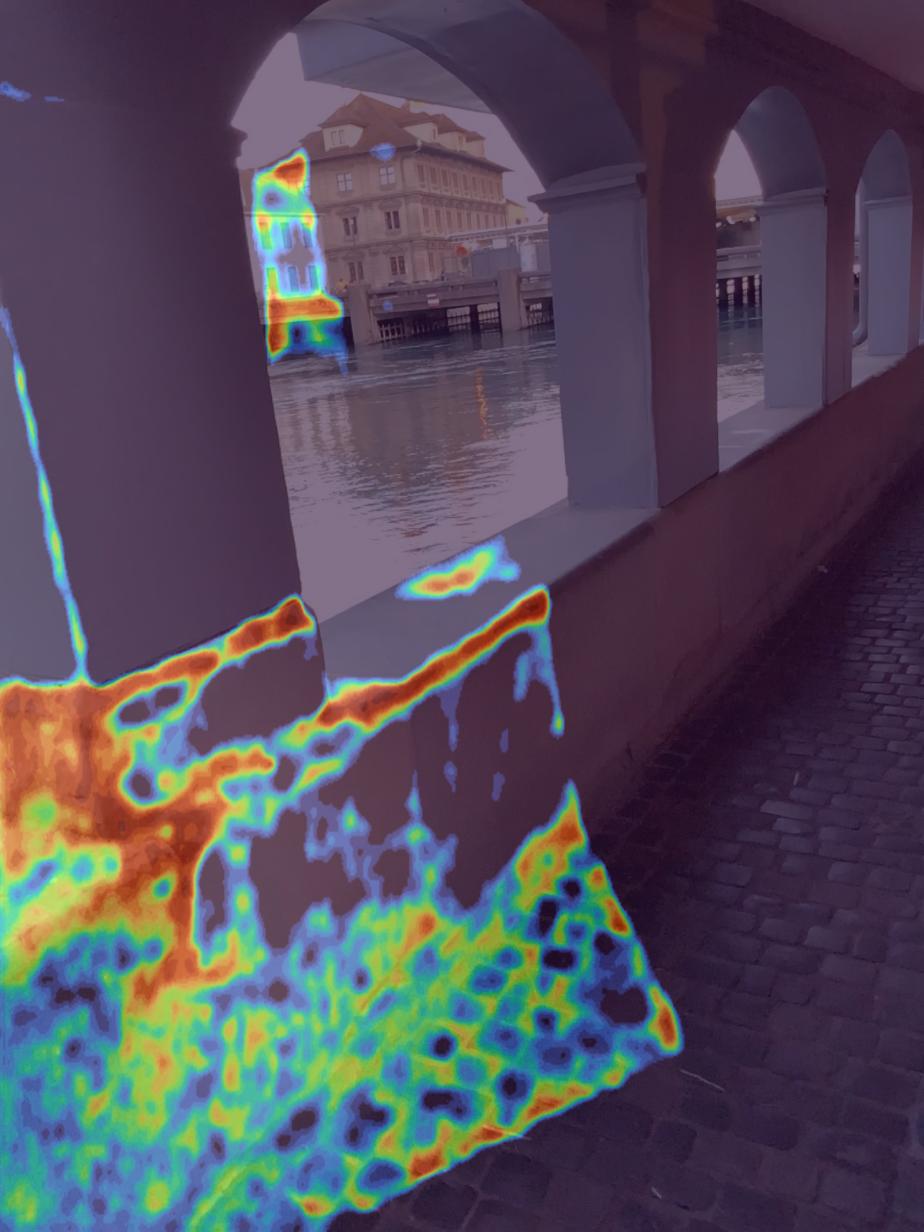}
    \caption{\textbf{Visualization of RoMa confidence} for two queries of the LaMAR~\cite{sarlin2022lamar} LIN dataset. The confidence values provide useful information to filter out unreliable regions such as moving people, vegetation, water, sky, enabling us to focus on stable textured regions. Colors encode high confidence (\emph{red}) and low confidence (\emph{blue}). Areas with dark color are below the threshold and are filtered out.
    }
    \label{fig:roma_confidence}
    \vspace{-0.15cm}
\end{figure*}

We visualize the confidence values of RoMa matching in \Fig\ref{fig:roma_confidence} for two queries. We observe that we can utilize the confidence to filter dynamic objects,~\eg, people, and uncertain regions,~\eg, vegetation and sky. 
Correspondences are established only for the colored areas with a sufficient confidence score. The confidences are also utilized as weights for the robust pose estimation with our GPU-accelerated LO-RANSAC~\cite{loransac}. 

\input{table/cambridge_matcher.tex}

\begin{figure*}[t]
    \centering
    \begin{subfigure}[b]{0.326\linewidth}
         \centering
         \includegraphics[width=\linewidth]{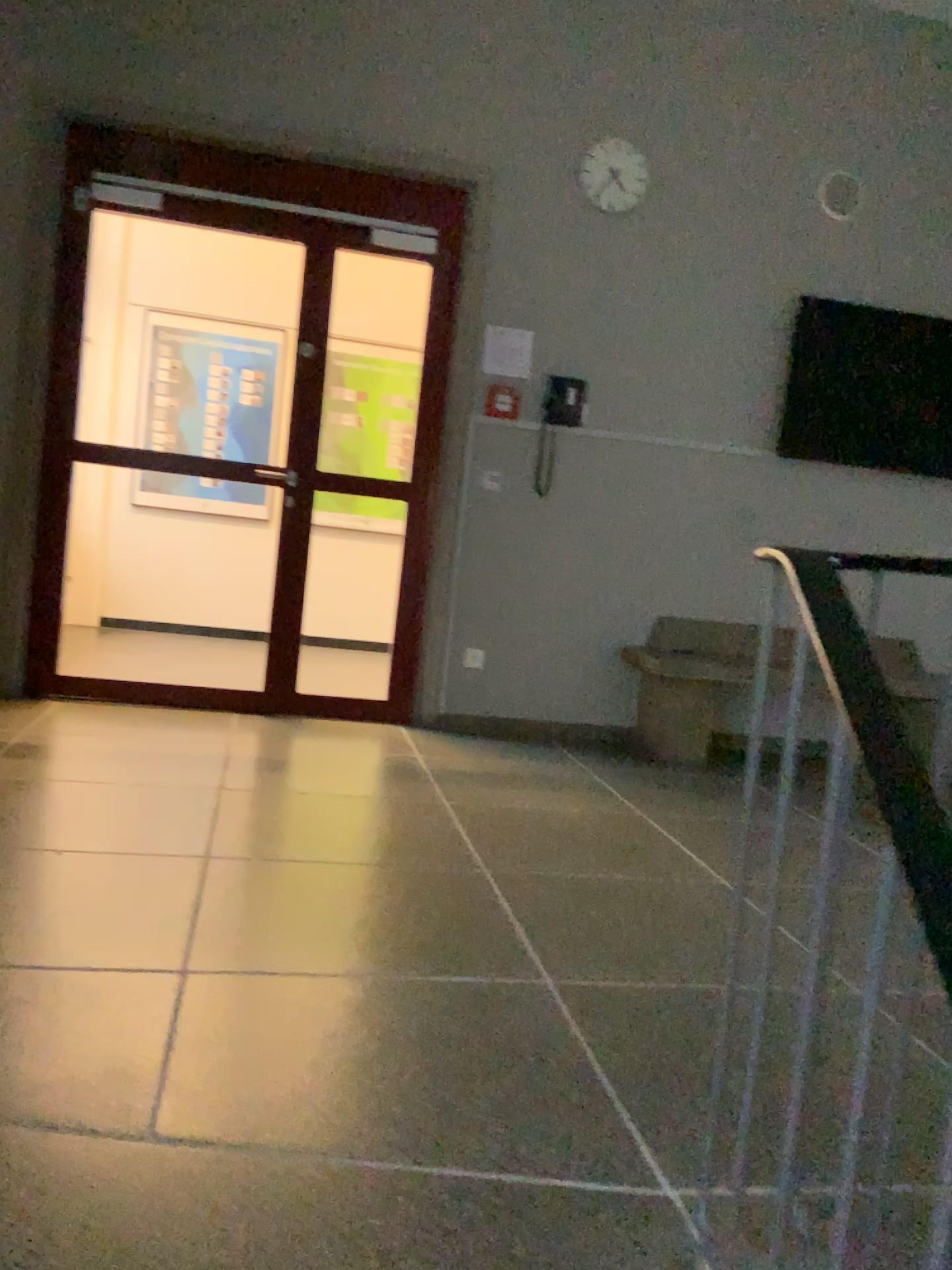}
         \caption{Query}
     \end{subfigure}
    \begin{subfigure}[b]{0.66\linewidth}
         \centering
         \includegraphics[width=\linewidth,height=0.66\linewidth]{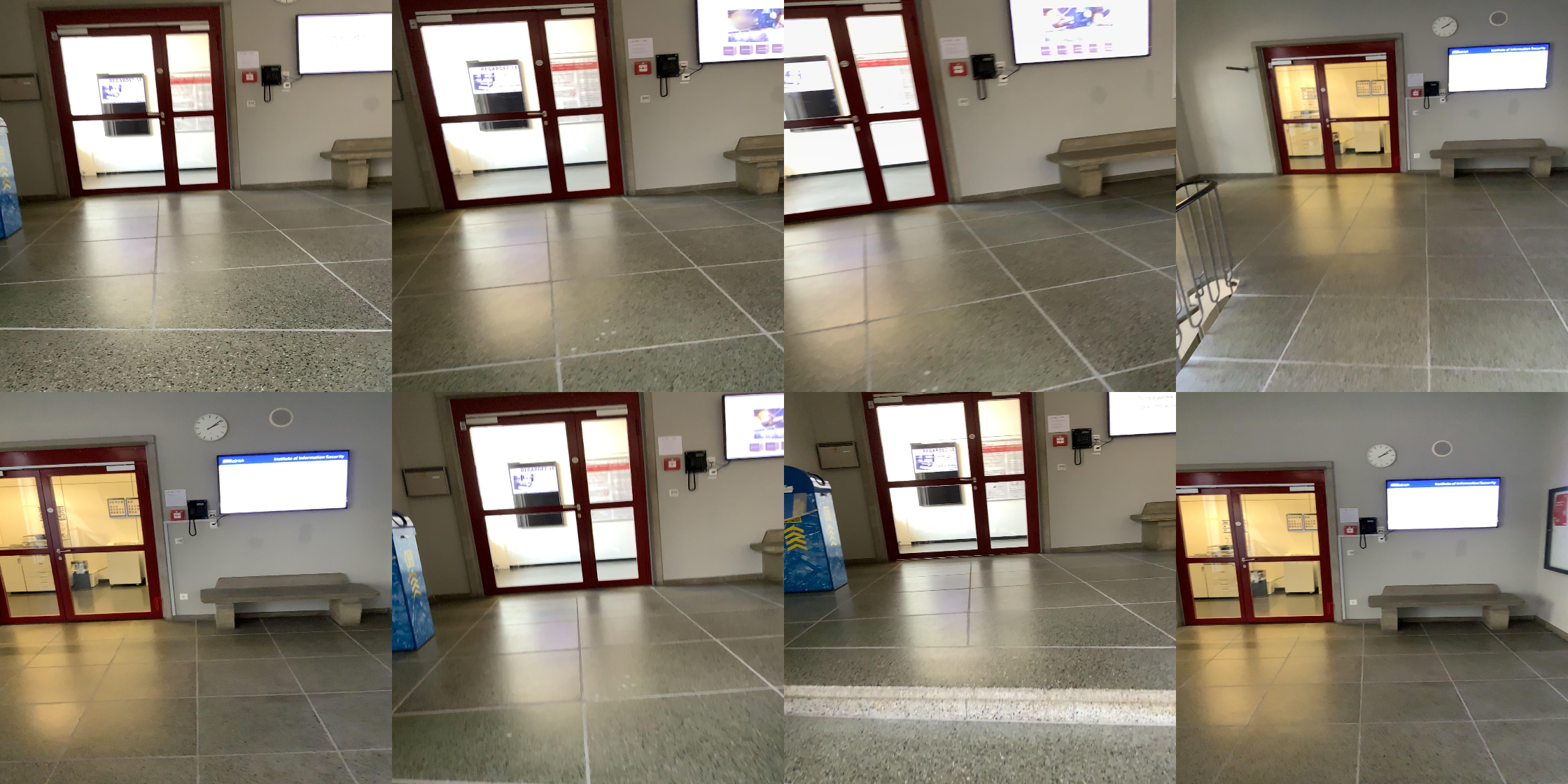}
         \caption{Megaloc~\cite{berton2025megaloc} retrieved images}
     \end{subfigure}
    
    \caption{\textbf{Typical failure case for \emph{ImLoc}} (LaMAR~\cite{sarlin2022lamar} data scene CAB). Due to repeated structures across the building floors the images retrieved by Megaloc~\cite{berton2025megaloc} are from wrong floors. Likewise, the RoMA matcher struggles to disambiguate repeated structures on different floors of the building.}
    \label{fig:floor_ambiguity}
    \vspace{-0.15cm}
\end{figure*}

\section{Limitations}
\subsection{Limitations of \emph{ImLoc}}
As shown in \Fig\ref{fig:floor_ambiguity}, we observe a common failure case of our method. 
Typically problems are induced from global ambiguities in the scenes. 
For instance, if multiple similar places exist in the map, such as different floors of a building. 
In this case, the retrieval method (Megaloc~\cite{berton2025megaloc}) may retrieve images from the wrong place, and sometimes the retrieved set might even lack any image from the correct place. 
The matcher (RoMa~\cite{edstedt2023roma}) usually is also not able to disambiguate the repeated structures, and cannot rescue the failure from retrieval. 
However, as shown in Fig~\ref{fig:aachen_matcher_compare}, recent large-scale feed-forward models like VGGT~\cite{wang2025vggt}, though not specifically trained to distinguish doppelgangers, already show strong potential.

\subsection{Limitations of the Pseudo Ground Truth}
\begin{figure}[t]
    \centering
    \includegraphics[width=0.49\linewidth]{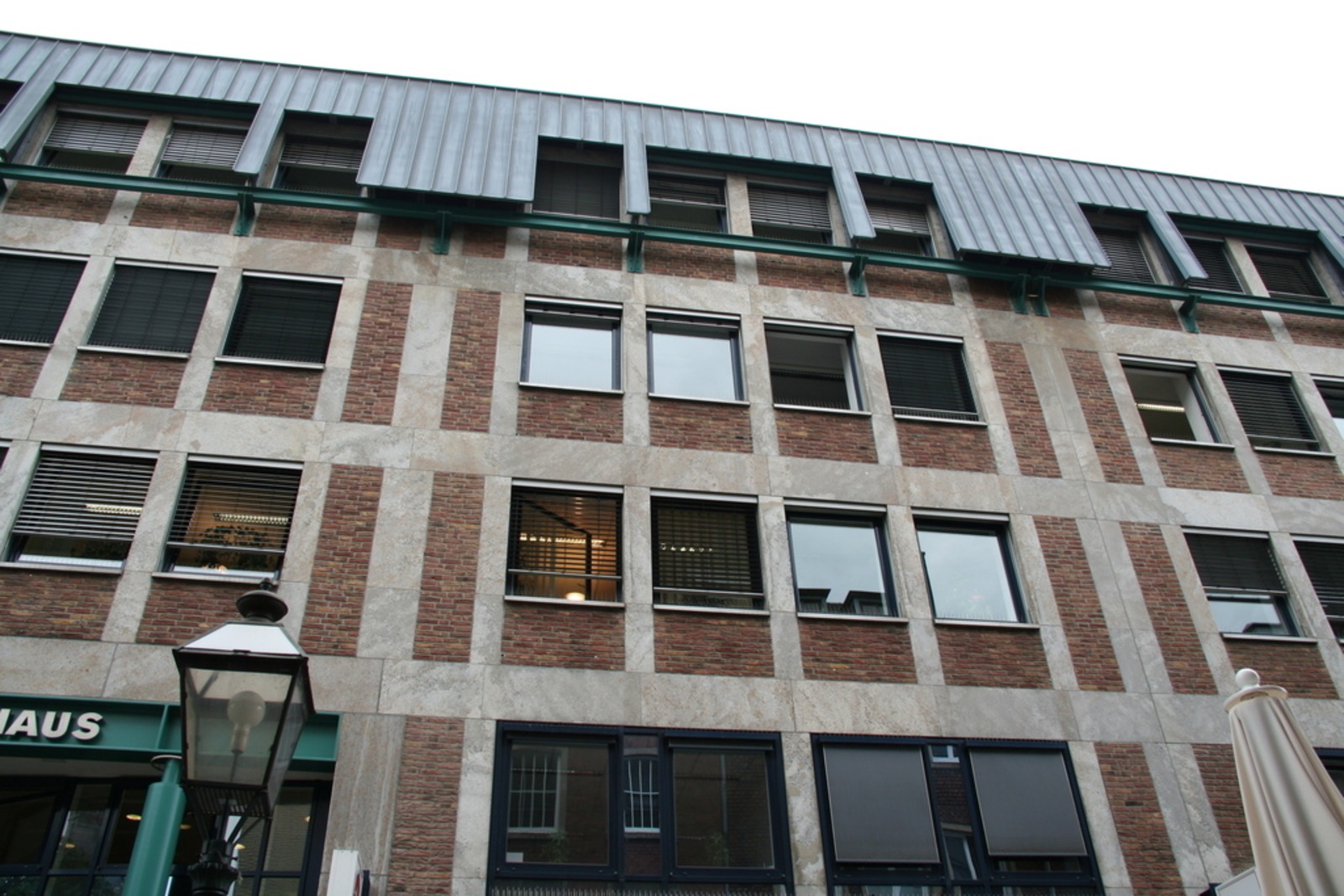}
    \includegraphics[width=0.49\linewidth]{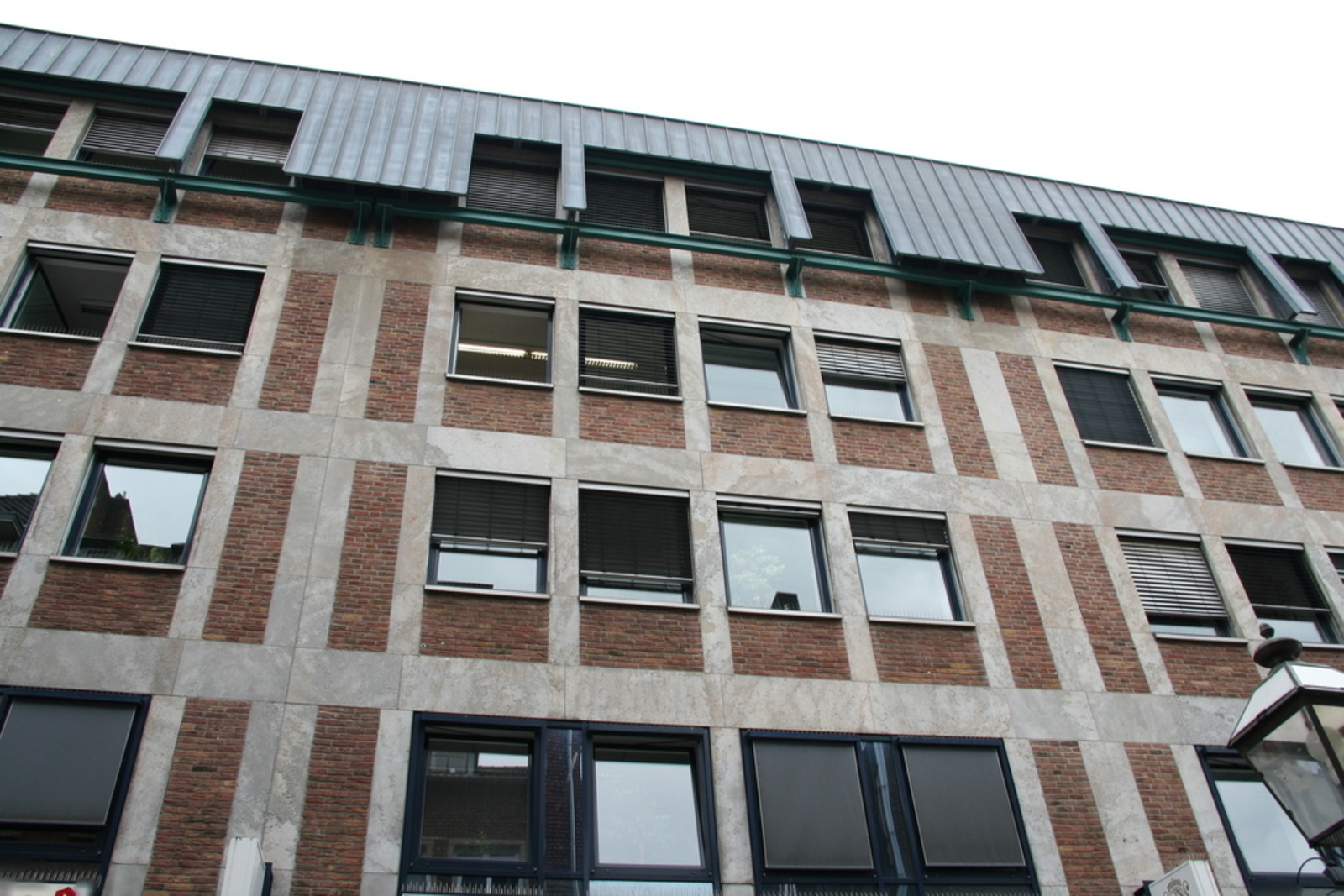}
    \includegraphics[width=0.49\linewidth]{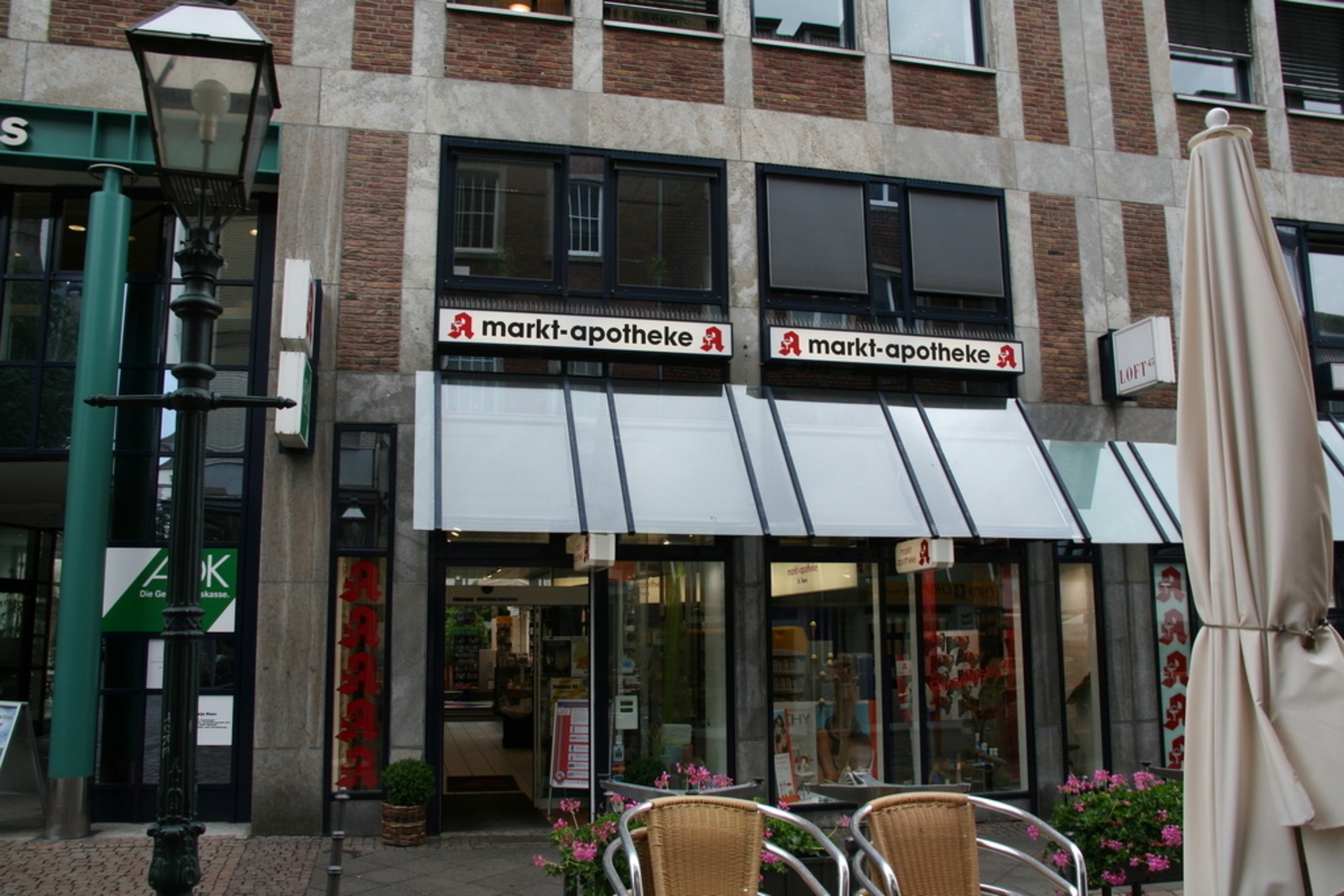}
    \includegraphics[width=0.49\linewidth]{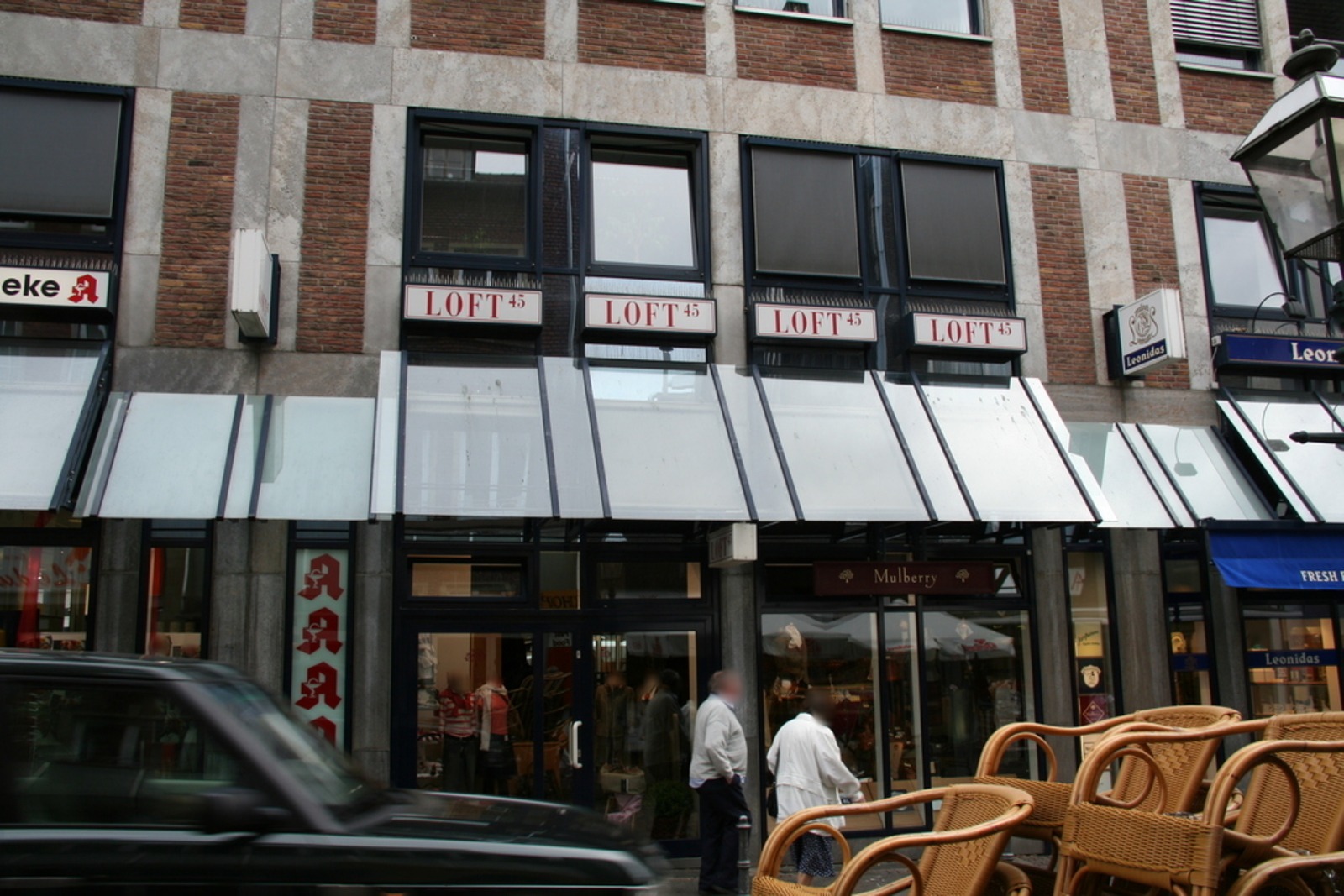}
    \caption{\textbf{Ground truth limitation}. Aachen Day Night~\cite{sattler2012bmvc, sattler2018cvpr} uses SfM to generate pseudo ground truth. This can produce wrong annotations when there is strong ambiguity. In this example, the two mapping images in the upper row have been wrongly labeled and were assigned a similar position, although they are actually observing different parts of the building (similar to Fig~\ref{fig:aachen_VGGT_dop}). This becomes more obvious by also considering the two images in the lower row.}
    \label{fig:aachen_gt}
    \vspace{-0.15cm}
\end{figure}

\begin{figure}[t]
    \centering
    \begin{subfigure}[b]{\linewidth}
         \centering
         \includegraphics[width=\linewidth,trim={0 12cm  0 12cm},clip]{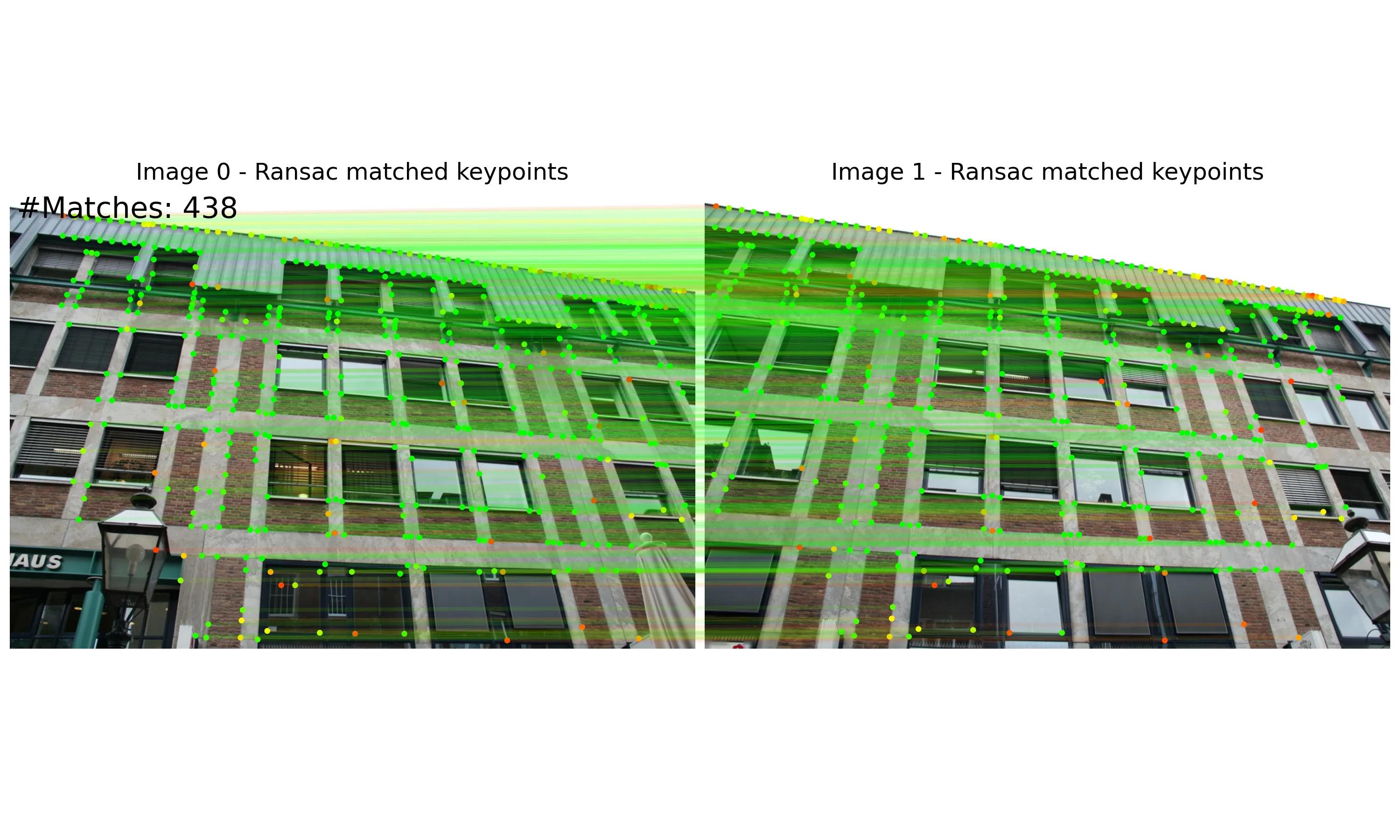}
         \caption{SP+LG~\cite{detone2018superpoint,lindenberger2023lightglue}}
     \end{subfigure}
     \begin{subfigure}[b]{\linewidth}
         \centering
         \includegraphics[width=\linewidth,trim={0 12cm  0 12cm},clip]{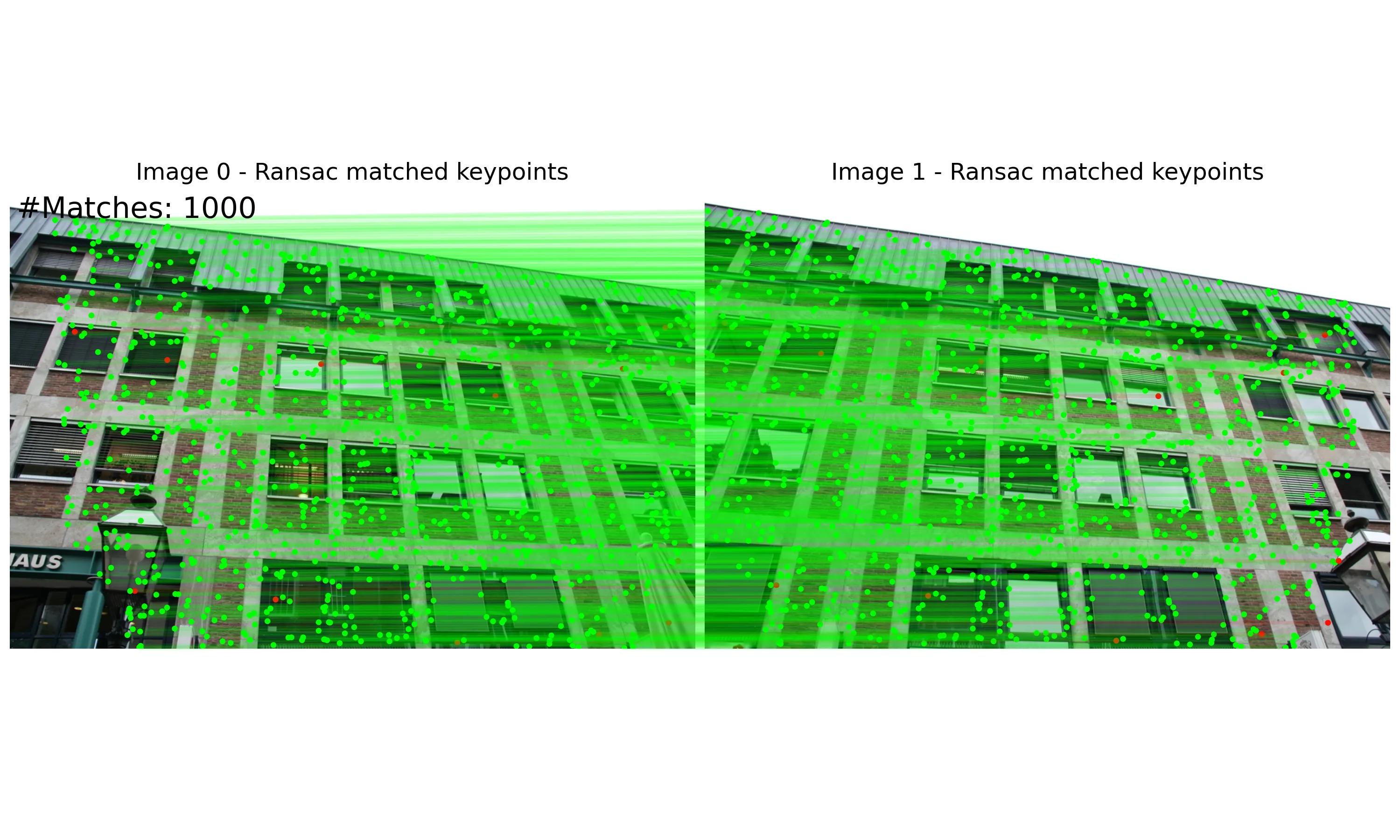}
         \caption{RoMa~\cite{edstedt2023roma}}
     \end{subfigure}
     \begin{subfigure}[b]{\linewidth}
         \centering
         \includegraphics[width=\linewidth,trim={0 12cm  0 12cm},clip]{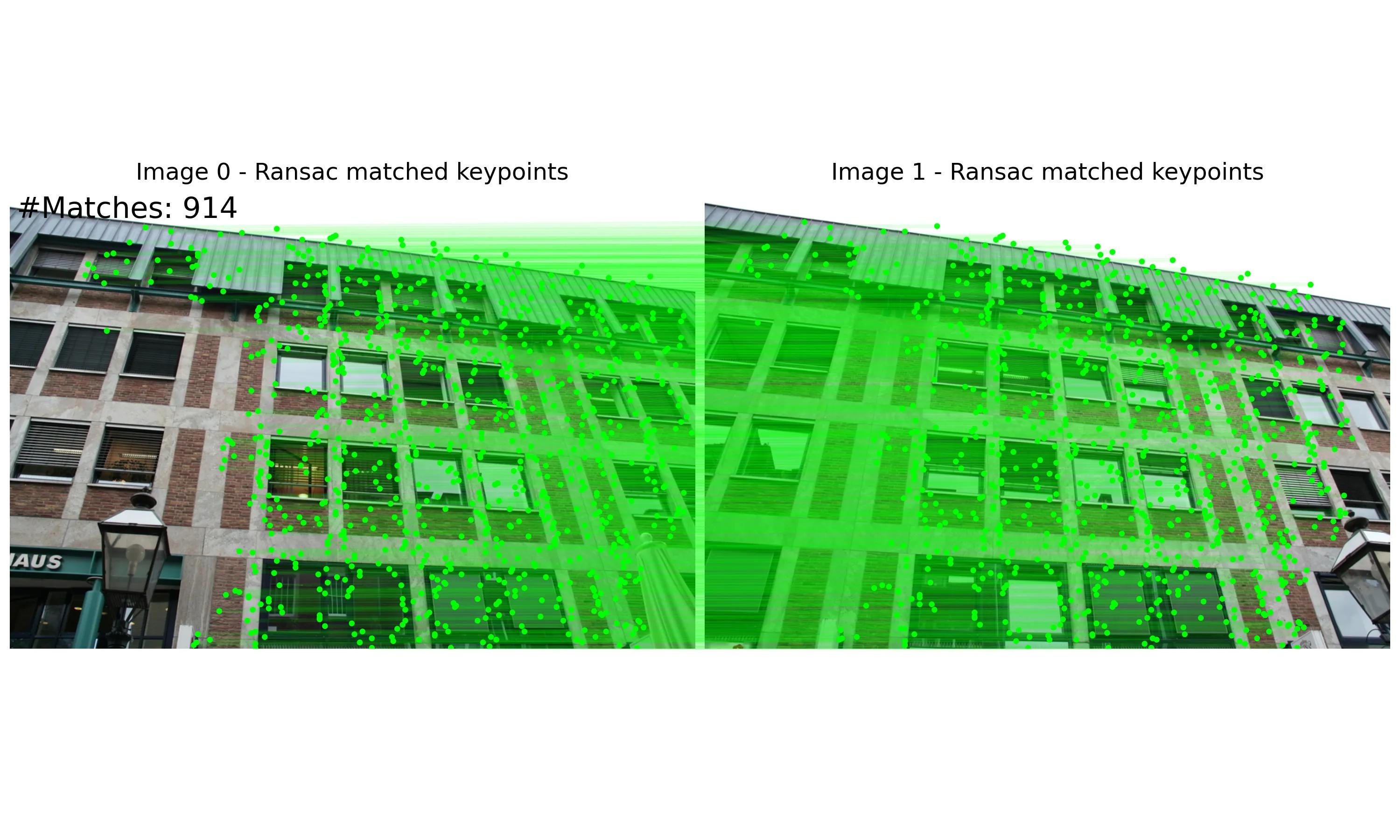}
         \caption{MASt3R~\cite{mast3r}}
     \end{subfigure}
     \begin{subfigure}[b]{\linewidth}
         \centering
         \includegraphics[width=\linewidth]{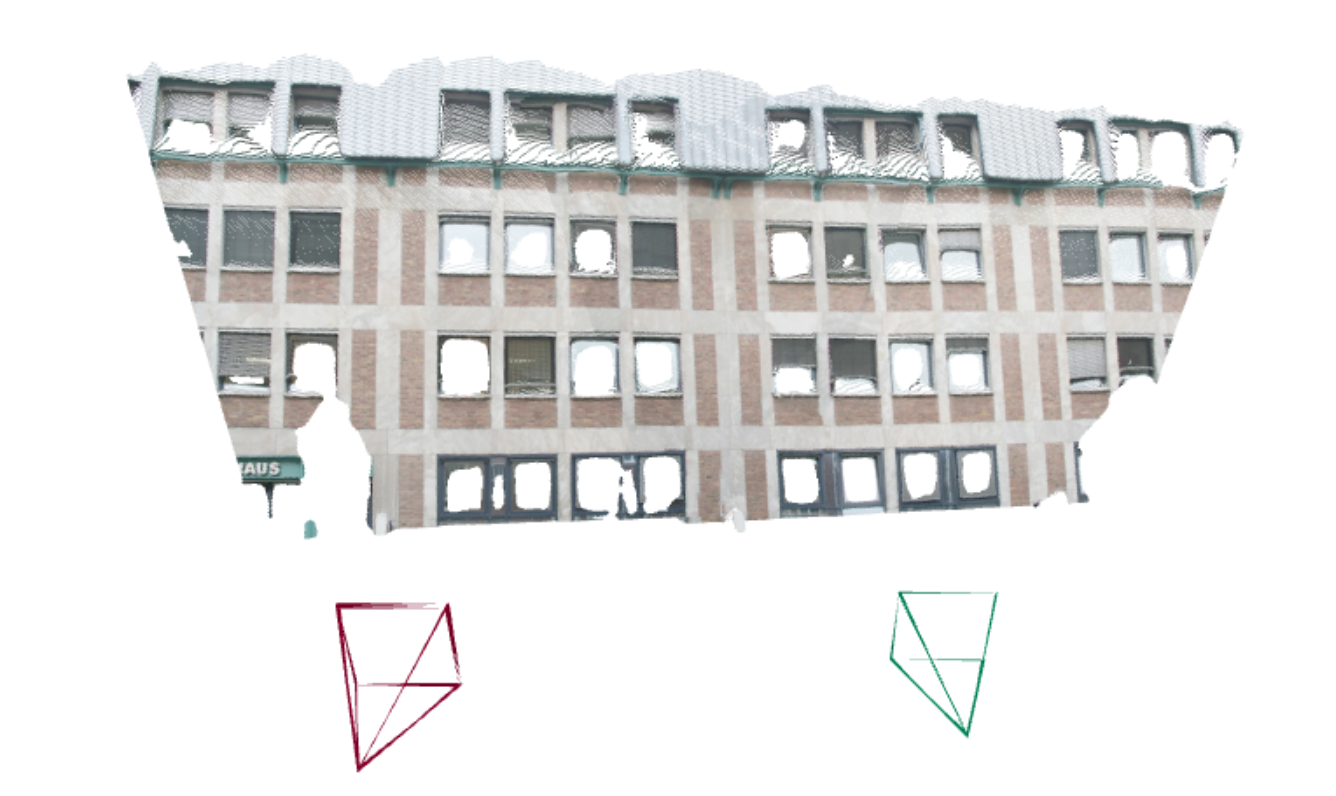}
         \caption{VGGT~\cite{wang2025vggt}}
         \label{fig:aachen_VGGT_dop}
     \end{subfigure}
    \caption{\textbf{Doppelgangers}. Doppelgangers can be challenging for many matchers~\cite{lindenberger2023lightglue,mast3r,edstedt2023roma}. However, recent large-scale feed-forward models like VGGT~\cite{wang2025vggt}, though not specifically trained to distinguish doppelgangers, already show strong potential. }
    \label{fig:aachen_matcher_compare}
    \vspace{-0.15cm}
\end{figure}

In addition to failures of our method at query time, we also observe that this kind of ambiguity may exist in the pseudo ground truth itself. 
Some datasets~\cite{kendall2015posenet,sattler2012bmvc, sattler2018cvpr} are using SfM to generate the pseudo ground truth, which may produce wrong annotations when there is strong ambiguity. 
As shown in \Fig\ref{fig:aachen_gt}, the top two mapping images are highly similar and have been wrongly labeled with similar position, however, they are actually looking at different parts of the building. This becomes more obvious, if we take the two images below into consideration. 
The actual poses are closer to the visualization of VGGT~\cite{wang2025vggt} reconstruction in \Fig\ref{fig:aachen_VGGT_dop}, which shows that these two images are looking at different parts of the wall. 
This kind of wrong labels occur in mapping images, and also may occur in the pseudo ground truth poses of query images. 
Without enough additional information to ensure the correctness of the annotations, these kinds of SfM pseudo ground truth methodologies appear somewhat limited. 
On the other hand, LaMAR~\cite{sarlin2022lamar} and Oxford Day \& Night~\cite{wang2025seeing} datasets acquire GT with additional information, including video sequence information, camera rig, IMU, LiDAR scan, which may be better choices to evaluate localization for challenges in the real-world.


%% file: table/lamar_hololens.tex
\begin{table*}
    \begin{center}
        \resizebox{\textwidth}{!}{
            \begin{tabular}{lccccccccccccccccccccccc}
                \toprule
                \multicolumn{1}{l}{\multirow{2}{*}{Methods}} &
                \multicolumn{2}{c}{CAB (val)}                &          &
                \multicolumn{2}{c}{HGE (val)}                &          &
                \multicolumn{2}{c}{LIN (val)}                &          &
                \multicolumn{2}{c}{Avg (val)}                &          &
                \multicolumn{2}{c}{CAB (test)}               &          &
                \multicolumn{2}{c}{HGE (test)}               &          &
                \multicolumn{2}{c}{LIN (test)}               &          &
                \multicolumn{2}{c}{Avg (test)}                                                                                                                                                                                                                    \\

                \cline{2-3} \cline{5-6} \cline{8-9} \cline{11-12} \cline{14-15} \cline{17-18} \cline{20-21} \cline{23-24}
                                                             & (1, 0.1) & (5, 1.0) &  & (1, 0.1) & (5, 1.0) &  & (1, 0.1) & (5, 1.0) &  & (1, 0.1) & (5, 1.0) &  & (1, 0.1) & (5, 1.0) &  & (1, 0.1) & (5, 1.0) &  & (1, 0.1) & (5, 1.0) &  & (1, 0.1) & (5, 1.0) \\
                \midrule
                Hloc (SP+SG)                                 & 81.07    & 93.35    &  & 77.86    & 83.16    &  & 95.63    & 98.96    &  & 84.86    & 91.82    &  & 78.67    & 90.23    &  & 74.80    & 85.59    &  & 86.55    & 97.18    &  & 80.01    & 91.00    \\
                Hloc (SP+LG)                                 & 79.88    & 93.86    &  & 77.01    & 82.48    &  & 95.08    & 98.61    &  & 83.99    & 91.65    &  & 78.06    & 89.85    &  & 74.21    & 84.25    &  & 85.94    & 97.15    &  & 79.40    & 90.42    \\
                \midrule
                Ours (RoMa)                                  & \textbf{86.79}    & \textbf{95.14}    &  & \textbf{78.55}    & \textbf{85.64}    &  & \textbf{95.70}    & \textbf{99.24}    &  & \textbf{87.01}    & \textbf{93.34}    &  & \textbf{80.53}    & \textbf{92.55}    &  & \textbf{78.27}    & \textbf{89.43}    &  & \textbf{88.90}    & \textbf{98.37}    &  & \textbf{82.57}    & \textbf{93.45}    \\

                \bottomrule
            \end{tabular}
        }
    \end{center}
    \caption{\textbf{Results on LaMAR dataset}, computed on each of the three scenes, for Hololens queries on validation set and submitted on the benchmark to obtain test set results. For each scene, we report the recall at (1°, 0.1m) and (5°, 1.0m), following the LaMAR paper~\cite{sarlin2022lamar}. We use 50 top-retrieved images for mapping and 10 top-retrieved images for localization using Megaloc~\cite{berton2025megaloc}. }
    \label{tab:lamar_hololens}
\end{table*}

%% file: table/cambridge_matcher.tex
\begin{table*}[]
    \centering
    \footnotesize
    \begin{tabular}{cllccccccc}
        \toprule
        \multicolumn{1}{l}{} &                                                                                      &                                                                           &                                                                      & \multicolumn{5}{c}{Cambridge Landmarks}                                                                                         \\
        \cmidrule(l){5-9}
        \multicolumn{1}{l}{} & \multirow{-2}{*}{\begin{tabular}[c]{@{}c@{}}Mapping\\ Matcher\end{tabular}}          & \multirow{-2}{*}{\begin{tabular}[c]{@{}c@{}}Query\\ Matcher\end{tabular}} & \multirow{-2}{*}{\begin{tabular}[c]{@{}c@{}}Map\\ Size\end{tabular}} & Court                                   & King's          & Hospital        & Shop           & St. Mary's
                             & \multirow{-2}{*}{\begin{tabular}[c]{@{}c@{}}Average \\ (cm / $^\circ$)\end{tabular}}                                                                                                                                                                                                                                                                                      \\
        \midrule
        \multirow{3}{*}{\rotatebox{90}{HLoc}}

                             & SP+LG~\cite{detone2018superpoint,lindenberger2023lightglue}                          & SP+LG~\cite{detone2018superpoint,lindenberger2023lightglue}               & $\sim$800MB                                                          & 18/0.1                                  & {\ul 12/0.2}     & 15/0.3          & \textbf{4/0.2} & \textbf{7/0.2} & {\ul 11/0.2}     \\
                             & LoFTR~\cite{sun2021loftr}                                                            & LoFTR~\cite{sun2021loftr}                                                 & $\sim$360MB                                                          & \textbf{16/0.1}                                  & 13/0.2          & 16/0.3  & \textbf{4/0.2}          & {\ul 8/0.2}         & 12/0.2          \\

                             & RoMa~\cite{edstedt2023roma}                                                          & RoMa~\cite{edstedt2023roma}                                               & $\sim$300MB                                                          & {\ul 17/0.1}                            & 15/0.2          & 17/0.3          & {\ul7/0.3}          & {\ul 8/0.2}    & 13/0.2          \\
        \midrule
        \multirow{3}{*}{\rotatebox{90}{Ours}}

                             & \multirow{3}{*}{RoMa~\cite{edstedt2023roma}}                                                                & SP+LG~\cite{detone2018superpoint,lindenberger2023lightglue}                                                                     & $\sim$90MB                                                           & {\ul17/0.1}                             & \textbf{11/0.2}          & {\ul 14/0.3}    & \textbf{4/0.2}          & \textbf{7/0.2}         & {\ul11/0.2}          \\
                             &                                                                                      & LoFTR~\cite{sun2021loftr}                                                                     & $\sim$90MB                                                           & {\ul17/0.1}                             & \textbf{11/0.2}          & \textbf{13/0.3} & \textbf{4/0.2}          & \textbf{7/0.2}          & \textbf{10/0.2}          \\
                             &                                                                                      & RoMa~\cite{edstedt2023roma}                                                                      & $\sim$90MB                                                           & \textbf{16/0.1}                         & \textbf{11/0.2} & {\ul 14/0.3}    & \textbf{4/0.2} & \textbf{7/0.2} & \textbf{10/0.2} \\

        \bottomrule
    \end{tabular}
    \caption{\textbf{Results on Cambridge Landmarks \cite{kendall2015posenet}.} We report median rotation and position errors. Best results are in \textbf{bold}, second best results are \underline{underlined}. For image retrieval, we use 10 images retrieved by NetVLAD~\cite{arandjelovic16netvlad} for all methods. Since we store the RGB images and dense geometry information, our method can flexibly switch to any sparse, semi-dense, dense matcher at query time.}
    \label{tab:cam_matcher_ablation}
\end{table*}